\documentclass{article}
\usepackage{arxiv,times}

\usepackage{amsmath,amsfonts,bm}

\def\eqref#1{equation~\ref{#1}}
\def\1{\bm{1}}

\DeclareMathAlphabet{\mathsfit}{\encodingdefault}{\sfdefault}{m}{sl}
\SetMathAlphabet{\mathsfit}{bold}{\encodingdefault}{\sfdefault}{bx}{n}

\usepackage{hyperref}
\usepackage{url}
\usepackage{graphicx}
\usepackage{svg}
\usepackage{algorithmicx}
\usepackage{algpseudocode}
\usepackage{tabularx}
\usepackage{xspace}
\usepackage[nopatch=footnote]{microtype}
\usepackage{siunitx}
\usepackage{mwe}
\usepackage[dvipsnames]{xcolor}
\usepackage{booktabs}
\usepackage{enumitem}
\usepackage[listings,most]{tcolorbox}
\usepackage{adjustbox}
\usepackage{colortbl}
\usepackage{multirow}
\usepackage{amssymb}
\usepackage{subcaption}
\usepackage{pifont}
\usepackage{algorithm}
\usepackage{wrapfig}
\usepackage{setspace}
\usepackage{tabularray}
\usepackage{array}
\usepackage{makecell}

\newcommand{\method}{\textsc{TimeInteract}\xspace}
\newcommand{\dataset}{\textsc{StreamTSI-34K}\xspace}
\newcommand{\logo}[1]{\raisebox{-0.25ex}{\includegraphics[height=1em]{#1}}\,}
\newcounter{annotationprompt}

\definecolor{PromptTitle}{HTML}{6C6875}
\definecolor{PromptBorder}{HTML}{AAA6B2}

\definecolor{headergreen}{RGB}{232,245,244}

\title{\method: Towards Real-Time Interactive Intelligence for Streaming Time Series}

\author{Sheng Pan$^{1}$, Yongli Gu$^{1}$, Yiqing Guo$^{2}$, Warren Jin$^{2}$, Bo Du$^{1}$, \textbf{Shirui Pan}$^{1}$\thanks{Corresponding authors.}, \textbf{Ming Jin}$^{1}$\footnotemark[1]\\
  $^{1}$Griffith University \\
  $^{2}$CSIRO \\
  \texttt{\small sheng.pan@griffithuni.edu.au}, \\
  \texttt{\small xgtx.weiyi@gmail.com} \\[2pt]
  \makebox[\textwidth][l]{\href{https://huggingface.co/TS-Interaction}
    {\logo{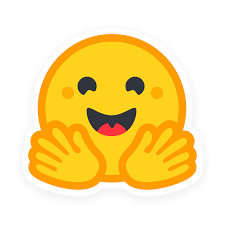}\,Dataset}
    \quad
    \href{https://github.com/shengpan-cpu/Timeinteract}
    {\logo{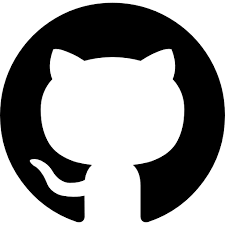}\,Code}
  }
}

\iclrfinalcopy
\begin{document}

\maketitle

\begin{abstract}
Real-world time series evolve continuously, with meaningful changes potentially emerging at any moment. However, existing time-series language models (TSLMs) remain inherently static. They either receive complete sequences for offline processing or alternate between streaming input and response generation, which prevents processing of new observations during interaction. We introduce a new regime, \textbf{Time-Series Interaction}: a model continuously perceives incoming time-series observations and user intent, autonomously decides when to remain silent or respond, and continues processing new observations during response generation. To realize this, we develop \textbf{\method} with three key designs: a \textit{dual-view streaming TS encoder} that captures local variations and historical dynamics, a \textit{response control mechanism} that learns when to trigger a response, and a \textit{decoupled streaming inference mechanism} that separates control from response generation to avoid blocking subsequent observations. We further formulate a hierarchy of interaction capabilities, progressing from Understanding to Adaptivity. Based on this hierarchy, we construct \textbf{\dataset}, a large-scale streaming TS interaction dataset with 34,588 episodes and 77,505 responses across synthetic and real-world time series in single- and multi-turn settings. Across all four interaction levels, \method consistently outperforms existing LLMs, VLMs, and TSLMs, with gains of up to 23.92 points on challenging tasks. It also improves response triggering while achieving near-zero stream stall and up to $2.15\times$ inference speedup.

\end{abstract}

\section{Introduction}

Streaming time series are prevalent across a wide range of real-world domains, including healthcare monitoring, industrial systems, financial markets, and the energy sector \citep{huang2026learning, helwig2015condition, jiang2024probabilistic, dong2026reinforcement,rogers2026improved}. In these scenarios, observations arrive incrementally as the underlying dynamics evolve \citep{wei2026evolving, li2026glucofm}, while users may issue new queries, set monitoring objectives, or adjust their goals throughout the stream \citep{yu2025sensorchat, kong2026timesage}. This requires continuously tracking the evolving data stream and providing timely responses as new observations arrive and user intent changes \citep{lau2025fast}.

\begin{figure}[ht]
    \centering
    \includegraphics[width=\textwidth]{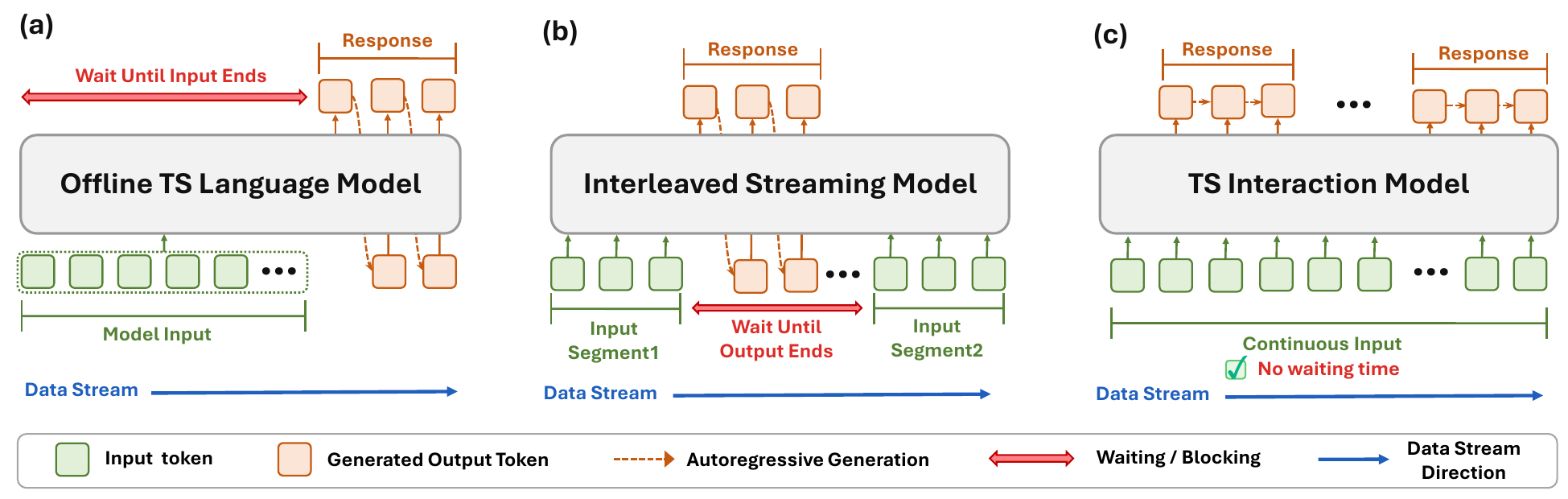}
    \caption{Overview of LLM-based paradigms for streaming time series. (a) Offline TS language models wait for the complete input before responding. (b) Interleaved streaming models alternate between input and output, causing input blocking during generation. (c) TS interaction models process incoming observations continuously while generating responses concurrently.}
    \vspace{-1mm}
    \label{fig1}
\end{figure}

In recent years, large language models (LLMs) have substantially broadened the scope of time-series analysis through their strong capabilities in language understanding, reasoning, and zero-shot generalization \citep{jin2024time, wang2024news,qiao2026s}. Beyond conventional tasks such as forecasting \citep{zhou2025time} and imputation \citep{guan2026timeomni}, LLM-based methods have enabled richer forms of time-series analysis, including question answering \citep{xie2024chatts} and temporal reasoning \citep{langer2025opentslm}. Despite these advances, unlocking the potential of LLMs for streaming TS interaction remains challenging. \textbf{\ding{182} Offline TS language models} (Fig.~\ref{fig1}(a)) operate on fixed input segments and respond only after the full sequence is available \citep{xie2024chatts,langer2025opentslm}. This works for low-frequency or post-hoc analysis, but introduces substantial latency on high-frequency streams. \textbf{\ding{183} Interleaved streaming models} (Fig.~\ref{fig1}(b)) reduce this latency by processing inputs incrementally and generating responses between successive input segments \citep{xie2026audio,wang2026think}. However, response generation still blocks subsequent observations, leaving additional latency during interaction. Although both designs can accommodate streaming inputs to some extent, they still fall short of real-time interaction because input reception and response generation remain coupled. \ding{184} Ideally, \textbf{TS interaction models} (Fig.~\ref{fig1}(c)) should continuously track the evolving stream, determine when a response is needed, and generate responses without interrupting subsequent inputs \citep{thinkingmachines2026interactionmodels}. This requires decoupling stream processing from response generation, allowing input and output to proceed concurrently.

However, several key gaps remain in the development of TS interaction models. First, \textbf{existing models do not naturally support streaming TS interaction} \citep{xie2024chatts,langer2025opentslm}. They lack key capabilities such as efficient stream encoding, response triggering, and concurrent generation. Second, \textbf{streaming TS interaction tasks lack a systematic formulation}, with different interaction forms and capability levels yet to be clearly defined. Third, \textbf{large-scale streaming TS interaction data remain scarce}. Most existing time-series QA datasets focus on isolated tasks or static question answering \citep{kong2025time, yu2026tsrbench}, with limited coverage of user queries, monitoring instructions, and multi-turn interactions over evolving streams.

To bridge these gaps, we develop \textbf{\method} specifically for streaming time-series interaction. It combines efficient stream encoding, response triggering, and concurrent generation, enabling the model to continuously follow incoming observations, determine when interaction is needed, and generate responses without interrupting the ongoing stream. We then formulate a hierarchy of streaming time-series interaction tasks that captures diverse interaction forms and difficulty levels, providing a structured basis for defining and evaluating interaction capabilities over evolving streams. Building on this task hierarchy, we construct \textbf{\dataset}, a large-scale dataset built from both synthetic and real-world time series, where interactions are grounded in the temporal evidence observed throughout the stream. It encompasses diverse streaming interaction scenarios across domains and difficulty levels, enabling models to learn appropriate interaction behaviors over continuously evolving time-series streams. Our main contributions are summarized as follows:

\begin{enumerate}
\setlength{\leftskip}{-2.1em}

    \item \textbf{A specialized model for streaming TS interaction.}
    We propose \textbf{\method} with three key designs: a \textbf{dual-view streaming TS encoder} for efficient stream encoding, a \textbf{response triggering mechanism} for deciding when to respond, and a \textbf{parallel control--response KV cache} that decouples stream processing from response generation. These designs decouple continuous input processing from response generation to support real-time TS interaction.

    \item \textbf{A systematic formulation of TS interaction tasks.} To capture the capability progression of TS interaction, we formulate a four-level hierarchy of TS interaction tasks with increasing interaction difficulty: \textbf{L1: Understanding} $\rightarrow$
    \textbf{L2: Persistence} $\rightarrow$
    \textbf{L3: Initiative} $\rightarrow$
    \textbf{L4: Adaptivity}. This hierarchy offers a unified way to organize and compare TS interaction capabilities at different levels of difficulty and enables models to be evaluated at different capability levels.

    \item \textbf{A large-scale streaming TS interaction dataset.} We construct \textbf{\dataset}, a large-scale dataset for streaming TS interaction. It combines controllable synthetic scenarios with diverse real-world time series and includes both \textbf{single-turn and multi-turn interactions}, capturing localized interaction behaviors as well as interaction continuity and cross-turn dependencies. These interactions cover \textbf{multiple interaction tasks} across the proposed capability hierarchy.

\end{enumerate}

\section{Related Work}

\textbf{LLM-Based Time-Series Analysis.} In recent years, LLMs have been increasingly adopted for time-series analysis \citep{zhang2024large,liu2025timecma,qin2026bridging}.
Initial efforts focused on transferring the knowledge and generalization ability of pretrained LLMs to conventional tasks such as forecasting and anomaly detection \citep{jia2024gpt4mts,jin2024time,sun2024test}.
The scope has since expanded beyond numerical tasks toward language-based time-series understanding \citep{cai2024timeseriesexam}.
Time-series language models (TSLMs) incorporate temporal signals through either textual inputs \citep{gruver2023large,zhou2023one}
or temporal representations interleaved with language \citep{xie2024chatts,langer2025opentslm,parker2025tsllm},
enabling description, question answering, and temporal reasoning. Recent work has moved toward more complex reasoning settings, including diverse reasoning capabilities and multi-turn analytical workflows \citep{guan2026timeomni,yu2026tsrbench,kong2026timesage}.
However, these methods largely operate on time-series context that is available in advance, rather than observations that continuously arrive and evolve during interaction.

\textbf{Streaming Multimodal Interaction.}
Real-time capability has become increasingly important for models processing continuously arriving multimodal signals \citep{lu2026survey,chen2025turn,hu2026tlive}. In audio, streaming ASR and spoken-dialogue systems reduce latency through incremental processing \citep{gao2022paraformer,xie2024mini,defossez2024moshi}. Recent work further explores richer interaction behaviors. Audio Interaction Model \citep{xie2026audio} unifies perception, response decisions, and generation, while JoyAI-VL-Interaction \citep{yao2026joyai} extends such interaction to live video by learning when to respond, remain silent, or delegate. These developments shift the focus from simply generating responses to deciding \emph{whether and when} to interact \citep{wang2024freeze,thinkingmachines2026interactionmodels,hu2026tlive}. However, existing interaction models remain centered on audio and visual modalities. To the best of our knowledge, we introduce the first interaction model specifically designed for streaming time series.

\section{\method}
\label{sec:method}

\textbf{Problem Definition.} Offline time-series language models receive a complete multivariate time series $\mathbf{X}\in\mathbb{R}^{L\times M}$ and generate a response to a given query or instruction as $r=f_{\theta}(\mathbf{X},\mathcal{E})$, where $L$ and $M$ denote the sequence length and number of variables, respectively. \method instead operates over a continuously evolving time-series stream $\mathbf{C}_{\leq t}$ with optional queries or instructions $\mathcal{E}_{\leq t}$, determining whether a response is required whenever new information becomes available:
\begin{equation}
    (d_t,r_t)
    =
    f_{\theta}
    \left(
        \mathbf{C}_{\leq t},
        \mathcal{E}_{\leq t},
        d_{<t},
        r_{<t}
    \right),
    \label{eq:streaming_interaction}
\end{equation}
where each incoming chunk is represented as $\mathbf{C}_t\in\mathbb{R}^{L_t\times M}$ and $\mathbf{C}_{\leq t}$ denotes the observations up to step $t$. $\mathcal{E}_{\leq t}$ denotes the user queries or instructions received by that time, while $d_{<t}$ and $r_{<t}$ record previous decisions and responses. The current decision $d_t\in\{\texttt{silent},\texttt{respond}\}$ determines the model's behavior: $d_t=\texttt{silent}$ continues stream processing without textual output, whereas $d_t=\texttt{respond}$ initiates response generation while new observations continue to arrive.

\subsection{Dual-View Streaming TS Encoder}
\label{sec:encoder}

\noindent\textbf{Dual-Reference Normalization.}
Offline time-series language models typically normalize a complete input sequence $\mathbf{X}\in\mathbb{R}^{L\times M}$ using statistics derived from the input itself \citep{langer2025opentslm}. In streaming settings, however, the incoming chunk $\mathbf{C}_t\in\mathbb{R}^{L_t\times M}$ contains both local temporal dynamics and changes relative to the accumulated historical stream \citep{lau2025fast, li2026glucofm}. A current view alone is therefore insufficient, and we construct two complementary views:
\begin{equation}
    \hat{\mathbf{C}}^{F}_t
    =
    \frac{
        \mathbf{C}_t-\boldsymbol{\mu}^{c}_t
    }{
        \boldsymbol{\sigma}^{c}_t+\epsilon
    },
    \qquad
    \hat{\mathbf{C}}^{H}_t
    =
    \frac{
        \mathbf{C}_t-\boldsymbol{\mu}^{h}_{t-1}
    }{
        \boldsymbol{\sigma}^{h}_{t-1}+\epsilon
    },
    \label{eq:dual_norm}
\end{equation}
where $\boldsymbol{\mu}^{c}_t,\boldsymbol{\sigma}^{c}_t\in\mathbb{R}^{M}$ are computed from the current chunk, while $\boldsymbol{\mu}^{h}_{t-1},\boldsymbol{\sigma}^{h}_{t-1}\in\mathbb{R}^{M}$ summarize the historical stream before step $t$. (1) The \emph{Fast View} $\hat{\mathbf{C}}^{F}_t$ emphasizes local variations within the current chunk. (2) The \emph{Historical View} $\hat{\mathbf{C}}^{H}_t$ captures how observations differ from the accumulated historical stream. Historical statistics are updated only after processing $\mathbf{C}_t$, preventing the current chunk from affecting its own historical view. The detailed update procedure is provided in Appendix~\ref{app:welford}.

\begin{figure}[ht]
    \centering
    \includegraphics[width=\textwidth]{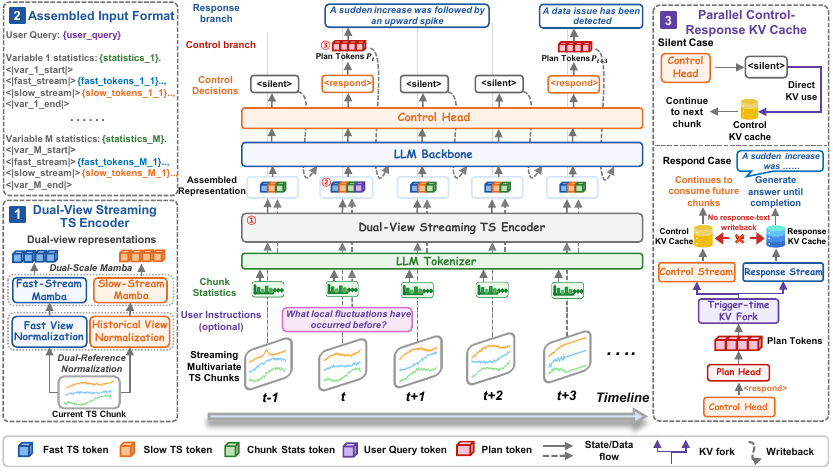}
    \caption{Overview of \method. (1) The dual-view streaming TS encoder captures local variations and historical dynamics. (2) Input assembly combines temporal tokens, chunk statistics, and user instructions. (3) KV forking supports concurrent stream processing and response generation.}
    \label{fig2}
\end{figure}

\noindent\textbf{Fast--Slow Stream Encoding.}
The normalized views $\hat{\mathbf{C}}^{F}_t$ and $\hat{\mathbf{C}}^{H}_t$ are first transformed into patch-level representations. Specifically, each variable is segmented along the temporal dimension into $J$ ordered patches and projected into an $H$-dimensional hidden space, yielding $\mathbf{E}^{F}_t,\mathbf{E}^{H}_t\in\mathbb{R}^{M\times J\times H}$. We then encode the two representations with independent Fast and Slow Mamba encoders:
\begin{equation}
    (\mathbf{Z}^{F}_t,\mathbf{S}^{F}_t)
    =
    \mathcal{M}_{F}(\mathbf{E}^{F}_t,\mathbf{S}^{F}_{t-1};\alpha_F),
    \qquad
    (\mathbf{Z}^{S}_t,\mathbf{S}^{S}_t)
    =
    \mathcal{M}_{S}(\mathbf{E}^{H}_t,\mathbf{S}^{S}_{t-1};\alpha_S).
    \label{eq:fast_slow}
\end{equation}
where $\mathcal{M}_{F}$ and $\mathcal{M}_{S}$ are independent Mamba encoders \citep{gu2023mamba}, and $\mathbf{S}^{F}_t$ and $\mathbf{S}^{S}_t$ are persistent SSM states carried across chunks. Their update rates satisfy $\alpha_F>\alpha_S$, allowing the Slow Stream to retain history longer while the Fast Stream responds more rapidly to recent changes. This persistent state enables incremental encoding without revisiting the full history. The resulting trajectories $\mathbf{Z}^{F}_t,\mathbf{Z}^{S}_t\in\mathbb{R}^{M\times J\times H}$ are further projected into the LLM hidden space:
\begin{equation}
    \mathbf{T}^{F}_t
    =
    \operatorname{RMSNorm}\!\left(\Pi_F(\mathbf{Z}^{F}_t)\right),
    \qquad
    \mathbf{T}^{S}_t
    =
    \operatorname{RMSNorm}\!\left(\Pi_S(\mathbf{Z}^{S}_t)\right),
    \label{eq:soft_tokens}
\end{equation}
where $\Pi_F,\Pi_S:\mathbb{R}^{H}\rightarrow\mathbb{R}^{D}$ are learned projections and $\mathbf{T}^{F}_t,\mathbf{T}^{S}_t\in\mathbb{R}^{M\times J\times D}$ denote the ordered Fast and Slow soft tokens. RMSNorm stabilizes temporal token scales via root mean square normalization and learned scaling. Each valid patch contributes one token per stream, and the tokens are inserted into the LLM context, where cross-variable dependencies are modeled through self-attention.

\subsection{Streaming Interaction Training}
\label{sec:training}

\noindent\textbf{Response Control and Latent Plans.} At each streaming step, \method forms a causal context $\mathcal{H}^{P}_t$ from the Fast and Slow tokens $\mathbf{T}^{F}_{\leq t},\mathbf{T}^{S}_{\leq t}$, user queries or instructions $\mathcal{E}_{\leq t}$, previous decisions $d_{<t}$, and Plans $\mathbf{P}_{<t}$ representing previous responses $r_{<t}$. A control head $G_{\mathrm{ctrl}}$ predicts $d_t\in\{\texttt{silent},\texttt{respond}\}$. Silent decisions continue stream processing without text, while response decisions activate a Plan head $G_{\mathrm{plan}}$ to produce  continuous Plan tokens $\mathbf{P}_t$ before decoding $r_t$:
\begin{gather}
    p_{\theta}(d_t\mid\mathcal{H}^{P}_t)
    =
    \operatorname{Softmax}\!\left(G_{\mathrm{ctrl}}(\mathbf{h}^{d}_t)\right),
    \label{eq:response_trigger} \\
    \mathbf{P}_t
    =
    G_{\mathrm{plan}}(\mathbf{h}^{p}_t)
    \in\mathbb{R}^{K\times D},
    \qquad \text{if } d_t=\texttt{respond},
    \label{eq:plan}
\end{gather}
where $\mathbf{h}^{d}_t$ and $\mathbf{h}^{p}_t$ are the LLM hidden states used by the control and Plan heads, respectively, $K$ is the number of Plan tokens, and $D$ is the LLM hidden dimension. We supervise response decisions with weighted cross-entropy $\mathcal{L}_{\mathrm{control}}$ to account for sparse response events. The Plan tokens $\mathbf{P}_t$ provide a compact representation of the intended response and remain in the control history to inform future decisions, allowing stream processing to continue without waiting for the complete response $r_t$.

\noindent\textbf{Full-to-Plan Distillation.}
To maintain a compact response history during inference, Plan tokens should preserve the information needed for subsequent interactions. We train the same language model under two aligned views: the \emph{Full View} $\mathcal{H}^{F}_t$ augments $\mathcal{H}^{P}_t$ with previous ground-truth responses $r^{*}_{<t}$, while the \emph{Plan View} retains only $\mathcal{H}^{P}_t$, excluding previous response text. Both views condition on the current Plan tokens $\mathbf{P}_t$ and the same target response prefix $r^{*}_{t,<j}$. We supervise Full-View response generation and distill its token distributions into the Plan View at every target response position:
\begin{gather}
    \mathcal{L}_{\mathrm{full}}
    =
    -\frac{1}{N_r}
    \sum_{t\in\mathcal{R}}\sum_{j=1}^{|r_t^*|}
    \log p_{\theta}\!\left(
        r^{*}_{t,j}
        \mid
        \mathcal{H}^{F}_t,
        \mathbf{P}_t,
        r^{*}_{t,<j}
    \right),
    \label{eq:full_loss} \\
    \mathcal{L}_{\mathrm{distill}}
    =
    \frac{\tau^2}{N_r}
    \sum_{t\in\mathcal{R}}\sum_{j=1}^{|r_t^*|}
    D_{\mathrm{KL}}\!\left[
        \operatorname{sg}(q^{F}_{t,j})
        \,\Vert\,
        q^{P}_{t,j}
    \right],
    \label{eq:distill_loss}
\end{gather}
where $\mathcal{R}$ indexes response steps and $N_r=\sum_{t\in\mathcal{R}}|r_t^*|$ is the total number of supervised response tokens. For view $v\in\{F,P\}$, $q^{v}_{t,j}=\operatorname{Softmax}(\mathbf{z}^{v}_{t,j}/\tau)$ denotes the vocabulary distribution obtained from logits $\mathbf{z}^{v}_{t,j}$ at temperature $\tau$. The operator $\operatorname{sg}(\cdot)$ stops gradients through the Full-View targets in distillation, while $\mathcal{L}_{\mathrm{full}}$ trains the shared model and Plan head. The distillation loss encourages the Plan tokens to preserve information required by future responses without retaining past replies.

\noindent\textbf{Three-Stage Training Pipeline.}
We adopt a three-stage training pipeline to learn the target sequence format, acquire the interaction mechanism, and scale training to diverse streaming settings. Training progressively introduces more complex interaction scenarios, in which previous Plan tokens inform subsequent response decisions and generation. At each stage, we optimize a weighted combination of the Response Control, Full-View generation, and Full-to-Plan distillation losses:
\begin{equation}
    \mathcal{L}^{(s)}
    =
    \lambda_c^{(s)}\mathcal{L}_{\mathrm{control}}
    +
    \lambda_f^{(s)}\mathcal{L}_{\mathrm{full}}
    +
    \lambda_d^{(s)}\mathcal{L}_{\mathrm{distill}},
    \qquad s\in\{1,2,3\},
    \label{eq:total_loss}
\end{equation}
where $s$ indexes the training stage and $\lambda_c^{(s)},\lambda_f^{(s)},\lambda_d^{(s)}$ are nonnegative loss weights. (1) \emph{Stage I: Temporal-Language Alignment.} we use simple interaction tasks in single-turn scenarios to learn the interaction format, training the streaming TS encoder and control and Plan heads with the LLM backbone frozen and distillation disabled ($\lambda_d^{(1)}=0$).
(2) \emph{Stage II: Interaction Mechanism Learning.} we train on all interaction tasks within single-turn scenarios, unfreeze the language model, and jointly optimize all three losses to learn response timing and how previous Plan tokens inform subsequent decisions. (3) \emph{Stage III: Large-Scale Streaming Interaction Training.} we extend joint training to the full data covering all interaction tasks in complex scenarios, including multi-turn sessions with mixed interaction forms and dependencies on earlier instructions, responses, and user feedback.

\subsection{Decoupled Streaming Inference via KV Forking}
\label{sec:inference}

Autoregressive response generation may span multiple time-series chunks
$\{\mathbf{C}_t\}$. In an interleaved setting, the model must finish
generating the current response before processing subsequent input,
causing response generation to block the incoming stream.
To remove this dependency, \method separates inference into a persistent
\emph{Control Stream} and temporary \emph{Response Streams}, using the
same language model with distinct KV caches.
The control cache $\mathbf{KV}^{\mathrm{ctrl}}$ follows the Plan View
$\mathcal{H}^{P}_t$: it accumulates temporal tokens, user instructions,
decisions, and Plan tokens, while the TS encoder carries its states
across input chunks.

\begin{wrapfigure}[19]{r}{0.5\textwidth}
    \footnotesize
    \hrule height 0.7pt
    \vskip 1.5pt

    \refstepcounter{algorithm}
    \noindent\textbf{Algorithm \thealgorithm\ Decoupled Streaming Inference}
    \label{alg:kv_forking}
    \label{alg:concurrent_inference}

    \vskip 1.5pt
    \hrule height 0.4pt
    \vskip 0.5pt

    \begin{algorithmic}[1]
        \algrenewcommand\algorithmicindent{1em}
        \algrenewcommand\algorithmicrequire{\textbf{Input:}}
        \algrenewcommand{\algorithmiccomment}[1]{\hfill\texttt{//}\ #1}

        \Require TS stream $\{\mathbf{C}_t\}$, optional user inputs $\{\mathcal{E}_t\}$
        \State Initialize $\mathbf{KV}^{\mathrm{ctrl}}$ and encoder states
        \State $\mathcal{Q}\gets\emptyset$; no active response

        \While{stream open or $\mathcal{Q}\neq\emptyset$}
            \If{a new chunk $\mathbf{C}_t$ is available}
                \State $\mathbf{B}_t \gets \Call{Encode}{\mathbf{C}_t}$
                \State $(d_t,\mathbf{KV}^{\mathrm{ctrl}})
                    \gets \Call{Control}{\mathbf{B}_t,\mathcal{E}_t,\mathbf{KV}^{\mathrm{ctrl}}}$

                \If{$d_t=\texttt{respond}$}
                    \State $(\mathbf{P}_t,\mathbf{KV}^{*}_t,\mathbf{h}^{r}_t)
                        \gets \Call{Plan}{\mathbf{KV}^{\mathrm{ctrl}}}$
                    \State $(\mathbf{KV}^{\mathrm{ctrl}},\mathbf{KV}^{\mathrm{resp}}_t)
                        \gets \Call{Fork}{\mathbf{KV}^{*}_t}$
                    \State $\mathbf{KV}^{\mathrm{ctrl}}
                        \gets \Call{FinishStep}{\mathbf{KV}^{\mathrm{ctrl}}}$
                    \State \Call{LaunchResponse}{$\mathcal{Q},\mathbf{KV}^{\mathrm{resp}}_t,\mathbf{h}^{r}_t$}
                \Else \Comment{\texttt{silent}}
                    \State $\mathbf{KV}^{\mathrm{ctrl}}
                        \gets \Call{FinishStep}{\mathbf{KV}^{\mathrm{ctrl}}}$
                \EndIf
            \EndIf

            \State \Call{CollectFinished}{$\mathcal{Q}$}
        \EndWhile
    \end{algorithmic}

    \vskip 0.5pt
    \hrule height 0.7pt

\end{wrapfigure}

At streaming step $t$, \textsc{Encode} operation encodes the incoming
chunk $\mathbf{C}_t$ into the temporal block $\mathbf{B}_t$, and
\textsc{Control} predicts and commits the binary response $d_t$.
User input is optional: $\mathcal{E}_t=\emptyset$ when no new query or
instruction arrives. A \texttt{silent} decision closes the control step
without creating a response request. For \texttt{respond},
\textsc{Plan} appends $\mathbf{P}_t$ and returns the cache
$\mathbf{KV}^{*}_t$ and hidden state $\mathbf{h}^{r}_t$ used to predict
the first response token. The cache is then forked into the continuing
control branch $\mathbf{KV}^{\mathrm{ctrl}}$ and a response branch
$\mathbf{KV}^{\mathrm{resp}}_t$. \textsc{FinishStep} closes the current
control step without modifying the response branch.
Subsequent cache updates remain isolated: new observations extend the
control history, while generated text extends only the response branch.
Each response stream is launched asynchronously by
\textsc{LaunchResponse} and tracked in $\mathcal{Q}$.
Once a response completes, its cache is released, while the corresponding
Plan tokens remain in the control history.
Algorithm~\ref{alg:kv_forking} summarizes the overall inference and response scheduling procedure.

\section{\dataset}
\label{sec:dataset}

\noindent\textbf{Overview.} \dataset is a large-scale dataset designed for streaming time-series interaction. The dataset comprises 34,588 interaction episodes and 77,505 annotated responses (1--10 per episode), drawing on both controllable synthetic signals and real-world time series from six domains. Across these diverse streams, we annotate four interaction tasks: \emph{Instant Query Answering (IQA)}, \emph{Persistent Instruction Following (PIF)}, \emph{Proactive Temporal Warning (PTW)}, and \emph{User-Guided Adaptation (UGA)}. These tasks are organized into single-turn episodes (56.9\%) and composed multi-turn interactions (43.1\%). Single-turn tasks ground response timing and content in temporal evidence and user intent, while multi-turn interactions capture cross-turn dependencies and interaction continuity.

\begin{figure}[ht]
    \centering
    \includegraphics[width=\textwidth]{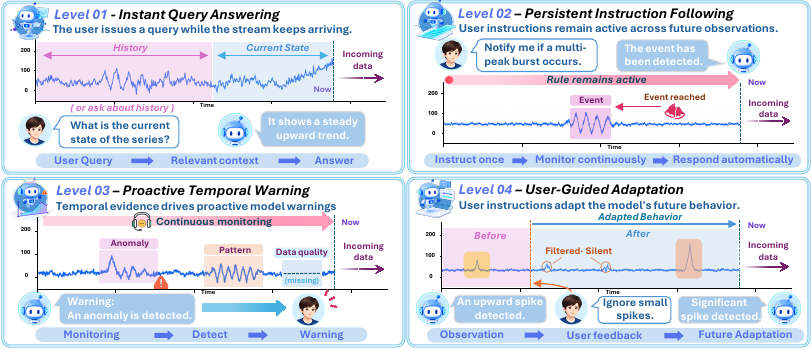}
    \caption{Representative cases of four streaming time-series interaction levels, from IQA to PIF, PTW, and UGA, corresponding to the capabilities of Understanding, Persistence, Initiative, and Adaptivity.}
    \label{fig3}
    \vspace{-3mm}
\end{figure}

\begin{figure}[ht]
    \centering
    \includegraphics[width=\textwidth]{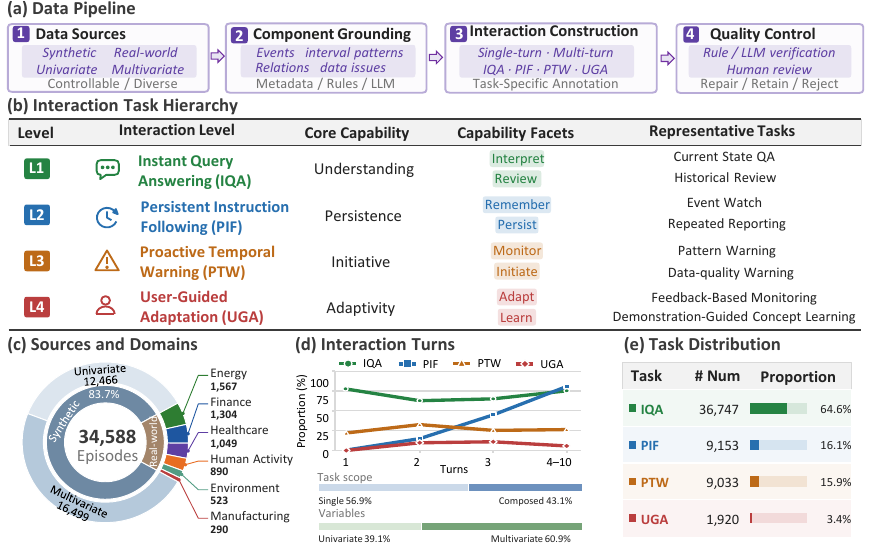}
    \caption{Construction and composition of \dataset. (a) Four-step data construction pipeline. (b) Interaction levels, capabilities, and representative tasks. (c) Source composition and real-world domain breakdown. (d) Task proportions by response count, alongside task scope and variable dimensionality. (e) Distribution of task instances, including tasks within composed episodes.}
    \vspace{-6mm}
    \label{fig4}
\end{figure}

\noindent\textbf{Data Sources.} We collect synthetic and real-world time series in both univariate and multivariate settings. Following ChatTS \citep{xie2024chatts}, synthetic data are generated from configurable trends, seasonal patterns, local events, and noise, with event attributes, temporal locations, and cross-variable relationships retained as metadata. We further manually curate real-world recordings from 12 datasets across six domains: energy, finance, healthcare, human activity, environment, and manufacturing. These data introduce diverse domain-specific dynamics and natural variability.

\noindent\textbf{Component Grounding.} We then identify interaction-relevant temporal components spanning local events, interval-level patterns, cross-variable relationships, and data-quality issues. These components range from isolated spikes and sustained upward or downward trends to inter-variable dependencies and corrupted signal segments. \ding{182} \emph{Synthetic data} use generation metadata to provide known temporal locations, whereas \ding{183} \emph{real-world data} are processed through LLM-assisted annotation to identify plausible interaction scenarios and localize the corresponding temporal evidence. The resulting components provide explicit temporal evidence for subsequent interaction data construction.

\noindent\textbf{Interaction Construction.} The grounded temporal evidence is converted into task-specific interactions for IQA, PIF, PTW, and UGA. Generation combines task-specific templates and rules with LLM-based generation to balance controllability and diversity. Each interaction specifies an optional user query or instruction, a response position, and the corresponding response text, with responses grounded only in observations available at or before the annotated position. Single-turn episodes are constructed to capture individual interaction behaviors, while multi-turn sessions compose multiple tasks so that earlier instructions, responses, and user feedback can influence subsequent interactions.

\noindent\textbf{Quality Control.} The generated interaction data first undergo LLM-based verification to ensure task consistency, appropriate response timing, and that each response relies only on observations available at the annotated position. Data that fail verification are regenerated using verifier feedback and re-evaluated, with this process repeated until they pass. Rule-based filtering then removes samples that violate task-specific requirements. High-confidence samples are retained directly, while low-confidence cases are manually reviewed to determine whether they should be accepted, corrected, or rejected. The complete data collection and construction procedure is provided in Appendix~\ref{app:dataset_details}.

\section{Experiment}
\label{sec:exp}

\noindent\textbf{Baselines.}
We compare against three categories of representative models. \textbf{General LLMs:} Qwen2.5-7B-Instruct \citep{qwen2025qwen25technicalreport}, Mistral-7B-Instruct-v0.3 \citep{jiang2023mistral7b}, and \mbox{Qwen3-14B~\citep{yang2025qwen3}}. \textbf{Vision-Language Models:} Qwen2.5-VL-7B \citep{bai2025qwen25vltechnicalreport} and InternVL3.5-8B \citep{wang2025internvl3}. \textbf{Time-Series Language Models:} ChatTS \citep{xie2024chatts} and TimeOmni-1 \citep{guan2026timeomni}. We report \method as the time-series interaction model.

\noindent\textbf{Evaluation.}
The evaluation covers all four interaction levels using task-specific metrics, including \textbf{\textit{Semantic Correctness (SC)}} for Understanding and \textbf{\textit{Instruction Fulfillment Rate (IFR)}} for Persistence, alongside metrics for Initiative and Adaptivity. Results are reported for both single-turn and multi-turn settings to capture interaction quality under different conversational contexts. Real-time efficiency is assessed using latency and responsiveness measures, including \textbf{\textit{Time to First Token (TTFT)}} and \textbf{\textit{Completion Latency}}, etc. Detailed metric definitions are provided in Appendix~\ref{app:evaluation_metrics}.

\subsection{Main Results}

We summarize the main results in Tables~\ref{tab:main_results} and~\ref{tab:triggering_results}.
\textbf{[Res.1] Strong interaction capability.}
\method ranks first across all four interaction levels in both settings, with the largest gains on UGA of +22.22 points in single-turn and +23.92 points in multi-turn interactions.
\textbf{[Res.2] Superior multi-turn performance.}
As the number of interaction turns increases, performance remains stable across all four levels (e.g., IQA: 67.34 $\rightarrow$ 64.56; PTW: 45.95 $\rightarrow$ 38.03), while consistently outperforming all baselines.
\textbf{[Res.3] Effective response triggering.}
\method achieves the highest F1 and NQ-F1 in both settings, indicating a better balance between responding and remaining silent. In multi-turn interactions, InternVL3.5-8B tends to over-trigger, reaching 98.24 recall but only 23.66 precision, whereas TimeOmni-1 is overly conservative and tends to suppress necessary responses.

\vspace{-4pt}

\begin{table}[ht]
\centering
\caption{
Performance on the four interaction levels of \dataset in single-turn and multi-turn settings.
\textbf{Native Streaming} indicates whether the model natively supports streaming TS data input.
}
\label{tab:main_results}

\setlength{\tabcolsep}{3pt}
\renewcommand{\arraystretch}{0.8}

\resizebox{\textwidth}{!}{\begin{tabular}{@{}l c c c cccc cccc@{}}
\toprule

\multirow{3.5}{*}{\textbf{Model}}
& \multirow{3.5}{*}{\textbf{Size}}
& \multirow{3.5}{*}{\makecell[c]{\textbf{Model}\\[-1pt]\textbf{Input}}}
& \multirow{3.5}{*}{\makecell[c]{\textbf{Native}\\[-1pt]\textbf{Streaming}}}
& \multicolumn{4}{c}{\textbf{Single-turn}}
& \multicolumn{4}{c}{\textbf{Multi-turn}} \\

\cmidrule(lr){5-8}
\cmidrule(lr){9-12}

& & & &
\makecell[c]{\textbf{IQA}\\[-1pt]{\scriptsize \textbf{\textit{(SC)}}}} &
\makecell[c]{\textbf{PIF}\\[-1pt]{\scriptsize \textbf{\textit{(IFR)}}}} &
\makecell[c]{\textbf{PTW}\\[-1pt]{\scriptsize \textbf{\textit{(CEHR)}}}} &
\makecell[c]{\textbf{UGA}\\[-1pt]{\scriptsize \textbf{\textit{(ASR)}}}} &
\makecell[c]{\textbf{IQA}\\[-1pt]{\scriptsize \textbf{\textit{(SC)}}}} &
\makecell[c]{\textbf{PIF}\\[-1pt]{\scriptsize \textbf{\textit{(IFR)}}}} &
\makecell[c]{\textbf{PTW}\\[-1pt]{\scriptsize \textbf{\textit{(CEHR)}}}} &
\makecell[c]{\textbf{UGA}\\[-1pt]{\scriptsize \textbf{\textit{(ASR)}}}} \\

\midrule

\multicolumn{12}
{@{}>{\columncolor{headergreen}[0pt][0pt]}l@{}}
{\textit{\textbf{General LLMs}}} \\

Qwen2.5-7B-Instruct
& 7B
& Text
& \ding{55}
& 44.14
& 3.45
& 2.70
& 0.00
& 48.30
& 2.52
& 1.41
& 0.00 \\

Mistral-7B-Instruct-v0.3
& 7B
& Text
& \ding{55}
& 38.06
& 0.00
& 5.41
& 0.00
& 35.13
& 0.00
& 1.41
& 0.00 \\

Qwen3-14B
& 14B
& Text
& \ding{55}
& \underline{55.63}
& 36.21
& 16.22
& 14.81
& \underline{55.95}
& 24.37
& 14.08
& \underline{13.04} \\

\multicolumn{12}
{@{}>{\columncolor{headergreen}[0pt][0pt]}l@{}}
{\textit{\textbf{Vision-Language Models}}} \\

Qwen2.5-VL-7B
& 7B
& Figure+Text
& \ding{55}
& 48.20
& 5.17
& 10.81
& 0.00
& 44.30
& 4.20
& 7.04
& 0.00 \\

InternVL3.5-8B
& 8B
& Figure+Text
& \ding{55}
& 55.18
& \underline{43.10}
& \underline{35.14}
& \underline{22.22}
& 53.16
& \underline{33.61}
& \underline{25.35}
& 8.70 \\

\multicolumn{12}
{@{}>{\columncolor{headergreen}[0pt][0pt]}l@{}}
{\textit{\textbf{Time-Series Language Models}}} \\

ChatTS
& 14B
& TS+Text
& \ding{55}
& 38.06
& 5.17
& 27.03
& 7.41
& 43.81
& 5.88
& 19.72
& 2.17 \\

TimeOmni-1
& 7B
& TS+Text
& \ding{55}
& 35.81
& 0.00
& 0.00
& 0.00
& 41.87
& 0.00
& 2.82
& 0.00 \\

\multicolumn{12}
{@{}>{\columncolor{headergreen}[0pt][0pt]}l@{}}
{\textit{\textbf{TS Interaction Model}}} \\

\textbf{\method}
& 4B
& Streaming TS+Text
& \ding{51}
& \textbf{67.34}
& \textbf{55.17}
& \textbf{45.95}
& \textbf{44.44}
& \textbf{64.56}
& \textbf{43.70}
& \textbf{38.03}
& \textbf{36.96} \\

\bottomrule
\end{tabular}}
\end{table}

\begin{table}[ht]
\centering
\caption{
Response-triggering performance on \dataset in single-turn and multi-turn settings.
\textbf{P}, \textbf{R}, \textbf{F1}, and \textbf{NQ-F1} denote triggering precision, recall, F1 score, and No-Query F1, respectively.}
\label{tab:triggering_results}
\Large
\setlength{\tabcolsep}{3pt}
\renewcommand{\arraystretch}{0.9}
\resizebox{\textwidth}{!}{\begin{tabular}{
@{}l c c c
*{8}{>{\centering\arraybackslash}m{3.3em}}
@{}
}
\toprule

\multirow{2.5}{*}{\textbf{Model}}
& \multirow{2.5}{*}{\textbf{Size}}
& \multirow{2.5}{*}{\makecell[c]{\textbf{Model}\\[-1pt]\textbf{Input}}}
& \multirow{2.5}{*}{\makecell[c]{\textbf{Native}\\[-1pt]\textbf{Streaming}}}
& \multicolumn{4}{c}{\textbf{Single-turn}}
& \multicolumn{4}{c}{\textbf{Multi-turn}} \\

\cmidrule(lr){5-8}
\cmidrule(lr){9-12}

& & & &
\textbf{P} &
\textbf{R} &
\textbf{F1} &
\mbox{\textbf{NQ-F1}} &
\textbf{P} &
\textbf{R} &
\textbf{F1} &
\mbox{\textbf{NQ-F1}} \\

\midrule

\multicolumn{12}
{@{}>{\columncolor{headergreen}[0pt][0pt]}l@{}}
{\textit{\textbf{General LLMs}}} \\

Qwen2.5-7B-Instruct
& 7B & Text & \ding{55}
& \underline{54.67} & 53.56 & \underline{54.11} & 5.30
& 83.60 & 62.87 & \underline{71.77} & 7.60 \\

Mistral-7B-Instruct-v0.3
& 7B & Text & \ding{55}
& 15.54 & 41.36 & 22.59 & 7.09
& 46.27 & 36.99 & 41.11 & 10.26 \\

Qwen3-14B
& 14B & Text & \ding{55}
& 15.81 & 74.92 & 26.11 & \underline{10.06}
& 33.88 & \underline{86.86} & 48.75 & 18.08 \\

\multicolumn{12}
{@{}>{\columncolor{headergreen}[0pt][0pt]}l@{}}
{\textit{\textbf{Vision-Language Models}}} \\

Qwen2.5-VL-7B
& 7B & Figure+Text & \ding{55}
& 14.91 & 49.49 & 22.92 & 5.67
& 48.55 & 58.94 & 53.24 & 9.80 \\

InternVL3.5-8B
& 8B & Figure+Text & \ding{55}
& 8.76 & \textbf{98.98} & 16.10 & 7.65
& 23.66 & \textbf{98.24} & 38.14 & 16.91 \\

\multicolumn{12}
{@{}>{\columncolor{headergreen}[0pt][0pt]}l@{}}
{\textit{\textbf{Time-Series Language Models}}} \\

ChatTS
& 14B & TS+Text & \ding{55}
& 7.83 & 58.31 & 13.80 & 7.43
& 22.00 & 44.04 & 29.35 & \underline{19.36} \\

TimeOmni-1
& 7B & TS+Text & \ding{55}
& 45.60 & 38.64 & 41.83 & 8.76
& \textbf{94.01} & 46.75 & 62.44 & 8.12 \\

\multicolumn{12}
{@{}>{\columncolor{headergreen}[0pt][0pt]}l@{}}
{\textit{\textbf{TS Interaction Model}}} \\

\textbf{\method}
& 4B & Streaming TS+Text & \ding{51}
& \textbf{87.46} & \underline{82.71} & \textbf{85.02} & \textbf{63.56}
& \underline{89.97} & 83.88 & \textbf{86.82} & \textbf{55.87} \\

\bottomrule
\end{tabular}}
\end{table}

\vspace{-6pt}

\subsection{Streaming Interaction Efficiency}

Beyond interaction quality, we assess real-time latency.
Table~\ref{tab:inference_efficiency} reports response latency,
stream processing efficiency, and speedup across response lengths.
\textbf{(1) Low first-token latency.}
\method exhibits similar TTFT to interleaved inference,
with negligible overhead from the Plan and control heads.
\textbf{(2) Higher efficiency for longer responses.}
By avoiding blocking between input processing and output generation,
\method achieves greater speedups for longer responses
(offline: $1.80\times \rightarrow 2.15\times$;
interleaved: $1.26\times \rightarrow 1.59\times$).
\textbf{(3) Efficient streaming interaction.}
Near-zero stall time and the lowest time per step across all groups
enable efficient real-time interaction.

\vspace{-6pt}
\begin{table}[ht]
\centering
\caption{
Inference efficiency across different response lengths, reported
in terms of response latency, stream processing, and overall speedup.
\textbf{Bold} values indicate the best latency results in each group.}
\label{tab:inference_efficiency}
\setlength{\tabcolsep}{3pt}
\renewcommand{\arraystretch}{0.9}
\large
\resizebox{\textwidth}{!}{\begin{tabular}{@{}c l cc cc c@{}}
\toprule
\multirow{2.5}{*}{\makecell[c]{\textbf{Response}\\[-1pt]\textbf{Length}}}
& \multirow{2.5}{*}{\textbf{Inference Strategy}}
& \multicolumn{2}{c}{\textbf{Response Latency}}
& \multicolumn{2}{c}{\textbf{Stream Processing}}
& \multicolumn{1}{c}{\textbf{Overall}} \\
\cmidrule(lr){3-4}
\cmidrule(lr){5-6}
\cmidrule(lr){7-7}
& &
\textbf{TTFT} $\downarrow$ (ms) &
\textbf{Completion Lat.} $\downarrow$ (ms) &
\textbf{Avg. Stall Time} $\downarrow$ (ms) &
\textbf{Time / Step} $\downarrow$ (ms) &
\textbf{Speedup} $\uparrow$ ($\times$) \\
\midrule

\multirow{3}{*}{$<50$ tokens}
& Offline
& 1368.05
& 2944.94
& ---
& 0.56
& 1.80 \\
& Interleaved
& 74.45
& 2065.39
& 856.50
& 0.39
& 1.26 \\
& \cellcolor{headergreen}\textbf{\method~(Ours)}
& \cellcolor{headergreen}\textbf{74.07}
& \cellcolor{headergreen}\textbf{1636.33}
& \cellcolor{headergreen}$\boldsymbol{\approx 0}$
& \cellcolor{headergreen}\textbf{0.31}
& \cellcolor{headergreen}--- \\
\midrule

\multirow{3}{*}{50--70 tokens}
& Offline
& 1578.12
& 3714.37
& ---
& 0.64
& 1.94 \\
& Interleaved
& 73.98
& 2703.60
& 1286.19
& 0.47
& 1.41 \\
& \cellcolor{headergreen}\textbf{\method~(Ours)}
& \cellcolor{headergreen}\textbf{72.30}
& \cellcolor{headergreen}\textbf{1913.09}
& \cellcolor{headergreen}$\boldsymbol{\approx 0}$
& \cellcolor{headergreen}\textbf{0.33}
& \cellcolor{headergreen}--- \\
\midrule

\multirow{3}{*}{$>70$ tokens}
& Offline
& 2041.13
& 5175.18
& ---
& 0.67
& 2.15 \\
& Interleaved
& \textbf{72.12}
& 3829.70
& 2044.13
& 0.50
& 1.59 \\
& \cellcolor{headergreen}\textbf{\method~(Ours)}
& \cellcolor{headergreen}73.32
& \cellcolor{headergreen}\textbf{2403.82}
& \cellcolor{headergreen}$\boldsymbol{\approx 0}$
& \cellcolor{headergreen}\textbf{0.31}
& \cellcolor{headergreen}--- \\
\bottomrule
\end{tabular}}
\end{table}

\begin{figure}[ht]
    \centering
    \includegraphics[width=\textwidth]{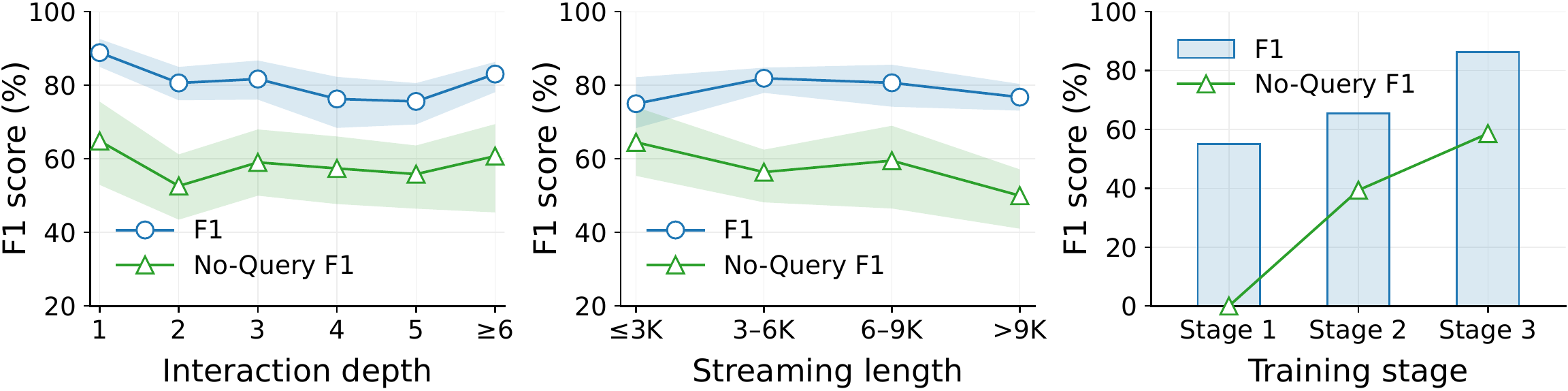}
    \caption{Analysis of interaction robustness and three-stage training. \textbf{Left and middle}: performance across interaction depths and stream lengths. \textbf{Right}: performance across three training stages.}
    \label{fig:fig5}
    \vspace{-4mm}
\end{figure}

\subsection{Additional Analysis}
\noindent
\begin{wrapfigure}{r}{0.48\textwidth}
\centering
\setlength{\parskip}{0pt}
\vspace{-4.5mm}
\includegraphics[width=\linewidth]{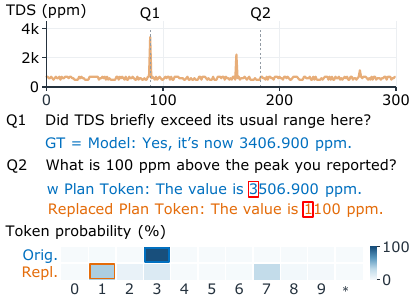}
\captionsetup{position=bottom,skip=4pt}
\caption{Visualization of Plan Token effectiveness in preserving
multi-turn information.}
\label{fig:representation}

\par\vspace{4pt}

\begingroup
\centering
\captionsetup{type=table,position=top,skip=3pt}
\caption{Ablation study of temporal representations and interaction mechanisms for triggering.}
\label{tab:ablation}
\footnotesize
\setlength{\tabcolsep}{3pt}
\renewcommand{\arraystretch}{0.95}

\begin{tabularx}{0.90\linewidth}{
    @{}>{\raggedright\arraybackslash}Xcc@{}
}
\toprule
\textbf{Variant} & \textbf{F1} & \textbf{NQ-F1} \\
\midrule

\multicolumn{3}
{@{}>{\columncolor{headergreen}[0pt][0pt]}l@{}}
{\textit{\textbf{Temporal Representation}}} \\

Repl. Slow TS token & 62.85 & 36.02 \\
Repl. Fast TS token & 71.16 & 45.05 \\

\midrule
\multicolumn{3}
{@{}>{\columncolor{headergreen}[0pt][0pt]}l@{}}
{\textit{\textbf{Interaction Mechanism}}} \\

Repl. Plan token & 76.13 & 48.53 \\

\midrule
\textbf{\method} & \textbf{86.30} & \textbf{58.61} \\
\bottomrule
\end{tabularx}
\par
\endgroup
\vspace{-6mm}
\end{wrapfigure}
\noindent\textbf{Effectiveness of Three-Stage Training.}
Figure~\ref{fig:fig5} shows that the three-stage training progressively improves the model's interaction capability. F1 increases from 55 $\rightarrow$ 65 $\rightarrow$ 86, while No-Query F1 rises substantially from 0 to 59. The gains from Stage I to Stage II suggest improved interaction understanding, while Stage III further boosts performance through large-scale streaming training.

\noindent\textbf{Information Retention in Plan Tokens.}
Figure~\ref{fig:representation} examines whether Plan Tokens preserve information across interaction turns. After identifying a peak of 3406.9 ppm, \method correctly answers the follow-up as $3506.9$ ppm using the Plan Token, despite the previous response text being absent. Replacing the Plan Token with a random embedding changes the answer to $1100$ ppm. The token distribution also assigns lower probability to the correct continuation. These results show that Plan Tokens effectively retain information from previous interactions.

\noindent\textbf{Ablation Study.}
We evaluate the contribution of each component by perturbing the embeddings of Slow TS tokens, Fast TS tokens, and Plan tokens. As shown in Table~\ref{tab:ablation}, all three perturbations reduce both F1 and NQ-F1. Perturbing Slow TS embeddings causes the largest degradation, indicating their greatest impact on response triggering. The performance drop caused by perturbing Plan embeddings further highlights the importance of retained interaction information for subsequent response decisions.

\noindent\textbf{Robustness to Longer Interactions.}
 Figure~\ref{fig:fig5} evaluates response triggering as interaction depth and stream length increase. Both metrics exhibit only moderate degradation as the interaction horizon grows, with F1 remaining consistently above 75 across all settings. This indicates that \method maintains stable response triggering over long interaction histories and streams.

\section{Conclusion}
\label{sec:con}

In this work, we introduce \textbf{Time-Series Interaction}, where models track evolving observations and user intent, decide when to respond, and continue processing inputs during generation. \method combines a dual-view streaming TS encoder, a response control mechanism, and decoupled streaming inference. We also establish a four-level interaction hierarchy and construct \dataset. Experiments show that \method achieves strong interaction quality and effective response triggering. Our work marks a step toward interactive intelligence for streaming time series.

\newpage

\bibliography{ref}
\bibliographystyle{main}

\newpage

\appendix

\section{Online Historical Statistics Update}
\label{app:welford}

We maintain the historical reference without storing or revisiting past observations. At each streaming step, we construct both normalized views before merging the current chunk into the historical statistics. This ordering avoids leakage from the current chunk into its historical reference.

\noindent\textbf{Reference Statistics and Normalization.}
Before processing $\mathbf{C}_t$, vectors $\mathbf{n},\boldsymbol{\mu},\mathbf{s}\in\mathbb{R}^{M}$ store historical counts, means, and centered sums of squares. They are initialized to zero. The current chunk yields statistics $\mathbf{n}^{c},\boldsymbol{\mu}^{c},\mathbf{s}^{c}$ from valid observations. For positive counts and valid positions, standard deviations and normalized views are computed element-wise with $\epsilon=10^{-5}$:
\begin{gather}
    \boldsymbol{\sigma}^{c}
    = \sqrt{\max\!\left(\frac{\mathbf{s}^{c}}{\mathbf{n}^{c}},\epsilon^2\right)},
    \qquad
    \boldsymbol{\sigma}
    = \sqrt{\max\!\left(\frac{\mathbf{s}}{\mathbf{n}},\epsilon^2\right)},
    \label{eq:reference_std} \\
    \hat{\mathbf{C}}^{F}_t
    = \frac{\mathbf{C}_t-\boldsymbol{\mu}^{c}}
           {\boldsymbol{\sigma}^{c}+\epsilon},
    \qquad
    \hat{\mathbf{C}}^{H}_t
    = \frac{\mathbf{C}_t-\boldsymbol{\mu}}
           {\boldsymbol{\sigma}+\epsilon}.
    \label{eq:appendix_dual_norm}
\end{gather}
The statistics are broadcast over time for each variable. The Fast View captures local shape within the current chunk, while the Historical View measures deviations from the preceding stream. For $\mathbf{n}=0$, both views use $\boldsymbol{\mu}^{c},\boldsymbol{\sigma}^{c}$. Padding values are excluded from statistics and zeroed in both views.

\noindent\textbf{Historical State Update.}
After forming both normalized views, we merge the current-chunk statistics into the historical state using a chunk-wise Welford update \citep{welford1962note}. Let $\boldsymbol{\delta}=\boldsymbol{\mu}^{c}-\boldsymbol{\mu}$ denote the mean difference, and let primes indicate updated quantities. For entries with $\mathbf{n}^{c}>0$, the historical statistics are updated element-wise to incorporate the current chunk as follows:
\begin{equation}
\begin{gathered}
    \mathbf{n}' = \mathbf{n}+\mathbf{n}^{c},
    \qquad
    \boldsymbol{\mu}' = \boldsymbol{\mu}
    +\frac{\mathbf{n}^{c}}{\mathbf{n}'}\odot\boldsymbol{\delta}, \\
    \mathbf{s}' = \mathbf{s}+\mathbf{s}^{c}
    +\frac{\mathbf{n}\odot\mathbf{n}^{c}}{\mathbf{n}'}
    \odot\boldsymbol{\delta}^{2}.
\end{gathered}
\label{eq:welford_update}
\end{equation}
Observation counts determine the relative weights of the historical and current-chunk means, while the correction in $\mathbf{s}'$ accounts for their difference. The merged statistics represent all valid observations accumulated up to the current chunk and provide the reference for normalizing the next chunk, leaving the current normalized views unchanged. Entries with $\mathbf{n}^{c}=0$ retain their previous statistics.

\section{Implementation Details}
\label{app:implementation_details}
\label{app:training_details}
\label{app:evaluation_metrics}

\noindent\textbf{Training Setup.} We initialize \method with Qwen3-4B-Instruct-2507 and follow the three-stage training pipeline. Stage I optimizes the streaming TS encoder, control and Plan heads while freezing the pretrained language model; Stages II and III jointly fine-tune all modules. Each subsequent stage starts from the preceding stage's model weights. All stages use AdamW with cosine learning-rate decay, 5\% warmup, BF16 precision, and DeepSpeed ZeRO-2. Training takes approximately five days on four H100 GPUs (94\,GB each). Table~\ref{tab:training_details} summarizes the training configuration.

\begin{table}[!htbp]
\centering
\caption{Hyperparameters and computational settings for the three-stage training pipeline.}
\label{tab:training_details}
\small
\setlength{\tabcolsep}{5pt}
\renewcommand{\arraystretch}{1.0}

\begin{tabularx}{\textwidth}{
    @{}l
    *{3}{>{\centering\arraybackslash}X}
    @{}
}
\toprule
\textbf{Parameter}
& \textbf{Stage I}
& \textbf{Stage II}
& \textbf{Stage III} \\
\midrule

\multicolumn{4}{@{}l}{\textit{Optimization}} \\

Epochs
& 1 & 1 & 3 \\

LLM learning rate
& Frozen & $1\times10^{-6}$ & $1\times10^{-5}$ \\

TS encoder and heads learning rate
& $2\times10^{-4}$ & $5\times10^{-5}$ & $1\times10^{-5}$ \\

Optimizer
& AdamW & AdamW & AdamW \\

Adam $(\beta_1,\beta_2)$
& $(0.9,\,0.999)$
& $(0.9,\,0.999)$
& $(0.9,\,0.999)$ \\

Adam $\epsilon$
& $10^{-8}$ & $10^{-8}$ & $10^{-8}$ \\

Learning-rate schedule
& Cosine decay & Cosine decay & Cosine decay \\

Warmup ratio
& 0.05 & 0.05 & 0.05 \\

Weight decay
& 0.01 & 0.01 & 0.01 \\

Maximum gradient norm
& 1.0 & 1.0 & 1.0 \\

\midrule
\multicolumn{4}{@{}l}{\textit{Training objectives}} \\

Loss weights $(\lambda_c,\lambda_f,\lambda_d)$
& $(0.2,1,0)$
& $(1,0.5,0.5)$
& $(1,0.5,0.5)$ \\

Control weights (\texttt{silent}:\texttt{respond})
& $1:5$ & $1:5$ & $1:5$ \\

Distillation temperature $\tau$
& -- & 2 & 2 \\

\midrule
\multicolumn{4}{@{}l}{\textit{Execution}} \\

GPUs
& \multicolumn{3}{c}{$4\times$ H100 (94\,GB each)} \\

Precision
& \multicolumn{3}{c}{BF16} \\

Sharding
& \multicolumn{3}{c}{DeepSpeed ZeRO-2} \\

Total training time (all stages)
& \multicolumn{3}{c}{5 days (approx.)} \\

\bottomrule
\end{tabularx}
\end{table}

\noindent\textbf{Interaction Quality.}
We evaluate the four interaction tasks using metrics aligned with their respective objectives. For IQA, \textbf{\textit{Semantic Correctness (SC)}} assigns each response a score $s_i^{\mathrm{SC}}\in[0,100]$, averaged over $N_{\mathrm{IQA}}$ instances. For PIF, \textbf{\textit{Instruction Fulfillment Rate (IFR)}} aggregates binary fulfillment indicators $f_i\in\{0,1\}$ over $N_{\mathrm{PIF}}$ instances. For PTW, \textbf{\textit{Correct Event Hit Rate (CEHR)}} measures the proportion of correctly warned events, denoted by $N_{\mathrm{hit}}$, among $N_{\mathrm{event}}$ annotated events. For UGA, \textbf{\textit{Adaptation Success Rate (ASR)}} measures the proportion of successful adaptations, denoted by $N_{\mathrm{success}}$, among $N_{\mathrm{UGA}}$ evaluated adaptations. These metrics are defined as follows:
\begin{equation}
\begin{aligned}
    \mathrm{SC}
    &= \frac{1}{N_{\mathrm{IQA}}}
       \sum_{i=1}^{N_{\mathrm{IQA}}}s_i^{\mathrm{SC}},
    \qquad\qquad
    &
    \mathrm{IFR}
    &= \frac{100}{N_{\mathrm{PIF}}}
       \sum_{i=1}^{N_{\mathrm{PIF}}}f_i,
    \\
    \mathrm{CEHR}
    &= 100\frac{N_{\mathrm{hit}}}{N_{\mathrm{event}}},
    \qquad\qquad
    &
    \mathrm{ASR}
    &= 100\frac{N_{\mathrm{success}}}{N_{\mathrm{UGA}}}.
\end{aligned}
\label{eq:interaction_metrics}
\end{equation}

\noindent\textbf{Interaction Efficiency.}
We evaluate real-time efficiency through response latency, stream continuity, and processing cost. \textbf{\textit{Time to First Token (TTFT)}} \citep{zhong2024distserve} and \textbf{\textit{Completion Latency}} \citep{agrawal2024taming} measure the delays from triggering to response initiation and completion, respectively. We denote the trigger time, first-token generation time, and response completion time by $t_{\mathrm{trig}}$, $t_{\mathrm{first}}$, and $t_{\mathrm{end}}$. \textbf{\textit{Average Stall Time}} measures how much response generation blocks incoming observations, averaging the blocked duration $\Delta t_i^{\mathrm{stall}}$ over $N_r$ triggered responses. \textbf{\textit{Time per Step}} \citep{sun2024llumnix} measures the average processing cost of an incoming step, using the total stream-processing time $T_{\mathrm{stream}}$ over $N_{\mathrm{step}}$ steps. Their mathematical definitions are as follows:
\begin{equation}
\begin{aligned}
    \mathrm{TTFT}
    &= t_{\mathrm{first}}-t_{\mathrm{trig}},
    \qquad\qquad
    &
    T_{\mathrm{comp}}
    &= t_{\mathrm{end}}-t_{\mathrm{trig}},
    \\
    \bar{T}_{\mathrm{stall}}
    &= \frac{1}{N_r}
       \sum_{i=1}^{N_r}\Delta t_i^{\mathrm{stall}},
    \qquad\qquad
    &
    T_{\mathrm{step}}
    &= \frac{T_{\mathrm{stream}}}{N_{\mathrm{step}}}.
\end{aligned}
\label{eq:efficiency_metrics}
\end{equation}
All four metrics are reported in milliseconds, with lower values indicating better efficiency. Near-zero stall time indicates that incoming observations can be processed with minimal interruption during response generation. We further report \textbf{\textit{Speedup}} to compare inference strategies, computed as $\mathrm{Speedup}=T_{\mathrm{comp}}^{\mathrm{baseline}}/T_{\mathrm{comp}}^{\mathrm{ours}}$, where $T_{\mathrm{comp}}^{\mathrm{ours}}$ denotes the completion latency of \method. This dimensionless ratio measures acceleration over the corresponding baseline.

\section{Dataset Details}
\label{app:dataset_details}

\subsection{Real-World Data Sources}
\label{app:real_data_sources}

The real-world portion of \dataset contains 5,623 interaction episodes constructed from 12 public datasets across six domains: energy, finance, healthcare, human activity, environment, and manufacturing. These sources cover heterogeneous temporal dynamics, sampling frequencies, signal modalities, and annotation schemes in both univariate and multivariate settings. For each source, we select variables suitable for streaming interaction and retain available event information for subsequent component grounding. Table~\ref{tab:real_data_sources} provides detailed information about the datasets.

\noindent\textbf{Energy.}
We use electricity-consumption signals from ENERTALK \citep{shin2019enertalk} and REFIT \citep{murray2017electrical}. For ENERTALK, we retain active-power channels from House~06, align independently timestamped measurements to a common temporal grid, and fill missing values. For REFIT, we combine dishwasher, washing-machine, and tumble-dryer loads from selected households into univariate sequences sampled at 8-second intervals \citep{guerrini2025time}.

\noindent\textbf{Finance.}
Financial sequences are drawn from BTCUSDT, which records the limit order book of Binance's Bitcoin--Tether perpetual contract. The source covers January~9--20, 2023, with approximately 3.73 million observations, providing high-frequency market recordings for interaction construction.

\noindent\textbf{Healthcare.}
The healthcare sources cover surgical motion, sleep physiology, and cardiac activity. For JIGSAWMaster, we use selected single-channel sequences from JIGSAWS \citep{gao2014jhu}, with labeled gesture examples and recurrence intervals supporting interactions around demonstrated and recurring motion patterns. From PhysioNet 2018 \citep{ghassemi2018you}, we select intervals from ten records and retain six synchronized channels. For MIT--BIH \citep{moody2001impact}, we extract ECG segments using cardiologist-reviewed beat annotations and AAMI beat-group labels.

\noindent\textbf{Human Activity.}
We combine wearable motion measurements, optical motion recordings, and physiological signals collected during physical activity. From PAMAP2 \citep{reiss2012introducing}, we retain six wrist and ankle accelerometer channels and construct contiguous 60--80-second windows. For Arm-CODA \citep{combettes2024arm}, we derive univariate displacement sequences from upper-limb motion recordings and use repeated segments to construct interactions involving movement recurrence. Pulse Transit Time PPG \citep{pollard2026physionet} provides physiological and motion recordings collected during sitting, walking, and running, covering different activity conditions.

\noindent\textbf{Environment.}
Environmental data come from the Macao Water Quality dataset \citep{gao2025water}, which monitors household tap-water conditions. Its four channels measure pH, turbidity, water temperature, and total dissolved solids, capturing changes in water quality.

\noindent\textbf{Manufacturing.}
We use industrial sensor recordings from SKAB \citep{skab} and the UCI Hydraulic dataset \citep{helwig2015condition}. SKAB records a laboratory water-circulation system under normal and abnormal conditions, with vibration, electrical, pressure, temperature, and flow measurements accompanied by anomaly and change-point annotations. The UCI Hydraulic dataset provides pressure, electrical power, flow, temperature, vibration, and efficiency measurements from repeated 60-second operating cycles, together with cycle-level condition labels.

\begin{table*}[t]
\centering
\caption{Statistics and composition of the real-world data in \dataset, covering 12 public datasets across six domains. The sources span diverse sampling frequencies, sequence lengths, and signal dimensionalities, providing a heterogeneous basis for constructing time-series interactions}
\label{tab:real_data_sources}

\footnotesize
\setlength{\tabcolsep}{2pt}

\resizebox{\textwidth}{!}{\begin{tabular}{@{}cccrrcrcrccc@{}}
\toprule
\multirow{2}{*}{\textbf{Dataset}} &
\multirow{2}{*}{\textbf{Dim}} &
\multirow{2}{*}{\textbf{Freq.}~\textbf{(Hz)}} &
\multicolumn{3}{c}{\textbf{Length}} &
\multicolumn{2}{c}{\textbf{Episodes}} &
\multicolumn{2}{c}{\textbf{Responses}} &
\multirow{2}{*}{\textbf{Source}} &
\multirow{2}{*}{\textbf{License}} \\
\cmidrule(lr){4-6}
\cmidrule(lr){7-8}
\cmidrule(lr){9-10}
& & &
\textbf{Min} &
\textbf{Max} &
\textbf{Avg} &
\textbf{Count} &
\textbf{Share (\%)} &
\textbf{Count} &
\textbf{Share (\%)} &
& \\
\midrule

\rowcolor{black!5}
\multicolumn{12}{l}{\textit{\textbf{Energy}}} \\

ENERTALK & 6 & 15
& 5,018 & 7,997 & 6,665.10
& 1,467 & 26.09 & 4,781 & 31.63
& \href{https://springernature.figshare.com/articles/dataset/House_06_in_ENERTALK_dataset/8123534}{Figshare}
& \href{https://creativecommons.org/publicdomain/zero/1.0/}{CC0 1.0} \\

REFIT & 1 & 0.13
& 5,000 & 7,989 & 5,763.90
& 100 & 1.78 & 200 & 1.32
& \href{https://pureportal.strath.ac.uk/en/datasets/refit-electrical-load-measurements-cleaned/}{Strathclyde}
& \href{https://creativecommons.org/licenses/by/4.0/}{CC BY 4.0} \\

\addlinespace[3pt]

\rowcolor{black!5}
\multicolumn{12}{l}{\textit{\textbf{Finance}}} \\

BTCUSDT & 4 & 4
& 5,000 & 5,000 & 5,000.00
& 1,304 & 23.19 & 3,904 & 25.83
& \href{https://www.kaggle.com/datasets/siavashraz/bitcoin-perpetualbtcusdtp-limit-order-book-data}{Kaggle}
& \href{https://www.kaggle.com/datasets/siavashraz/bitcoin-perpetualbtcusdtp-limit-order-book-data}{MIT}$^{a}$ \\

\addlinespace[3pt]

\rowcolor{black!5}
\multicolumn{12}{l}{\textit{\textbf{Healthcare}}} \\

JIGSAWMaster & 1 & 30
& 3,000 & 3,000 & 3,000.00
& 245 & 4.36 & 769 & 5.09
& \href{https://cirl.lcsr.jhu.edu/research/hmm/datasets/jigsaws_release/}{JHU--ISI}
& \href{https://cirl.lcsr.jhu.edu/research/hmm/datasets/jigsaws_release/}{Academic}$^{b}$ \\

PhysioNet 2018 & 6 & 200
& 8,000 & 8,000 & 8,000.00
& 459 & 8.16 & 715 & 4.73
& \href{https://physionet.org/content/challenge-2018/1.0.0/}{PhysioNet}
& \href{https://physionet.org/content/challenge-2018/view-license/1.0.0/}{ODC-By 1.0} \\

MIT--BIH & 2 & 360
& 2,920 & 16,675 & 6,703.00
& 345 & 6.14 & 1,019 & 6.74
& \href{https://physionet.org/content/mitdb/1.0.0/}{PhysioNet}
& \href{https://physionet.org/content/mitdb/view-license/1.0.0/}{ODC-By 1.0} \\

\addlinespace[3pt]

\rowcolor{black!5}
\multicolumn{12}{l}{\textit{\textbf{Human Activity}}} \\

PAMAP2 & 6 & 100
& 6,000 & 8,000 & 7,381.80
& 598 & 10.63 & 1,196 & 7.91
& \href{https://archive.ics.uci.edu/dataset/231/pamap2+physical+activity+monitoring}{UCI}
& \href{https://creativecommons.org/licenses/by/4.0/}{CC BY 4.0} \\

Arm-CODA & 1 & 100
& 6,603 & 9,420 & 8,047.40
& 192 & 3.41 & 384 & 2.54
& \href{https://www.ipol.im/pub/art/2024/494/}{IPOL}
& \href{https://creativecommons.org/licenses/by-nc-sa/3.0/}{CC BY-NC-SA 3.0} \\

PTT PPG & 1 & 500
& 1,061 & 3,296 & 2,169.30
& 100 & 1.78 & 350 & 2.32
& \href{https://physionet.org/content/pulse-transit-time-ppg/1.1.0/}{PhysioNet}
& \href{https://physionet.org/content/pulse-transit-time-ppg/view-license/1.1.0/}{ODbL 1.0} \\

\addlinespace[3pt]

\rowcolor{black!5}
\multicolumn{12}{l}{\textit{\textbf{Environment}}} \\

Macao Water & 4 & 0.20
& 6,000 & 6,000 & 6,000.00
& 523 & 9.30 & 1,109 & 7.34
& \href{https://github.com/PriGaoJiawei/Macao-water-dataset}{GitHub}
& \href{https://github.com/PriGaoJiawei/Macao-water-dataset}{Not stated} \\

\addlinespace[3pt]

\rowcolor{black!5}
\multicolumn{12}{l}{\textit{\textbf{Manufacturing}}} \\

SKAB & 5 & 1
& 745 & 1,327 & 1,133.10
& 30 & 0.53 & 30 & 0.20
& \href{https://github.com/waico/SKAB}{GitHub}
& \href{https://github.com/waico/SKAB/blob/master/LICENSE}{GPL-3.0}$^{c}$ \\

UCI Hydraulic & 3 & 100
& 6,000 & 24,000 & 14,571.40
& 260 & 4.62 & 659 & 4.36
& \href{https://archive.ics.uci.edu/dataset/447/condition+monitoring+of+hydraulic+systems}{UCI}
& \href{https://creativecommons.org/licenses/by/4.0/}{CC BY 4.0} \\

\bottomrule
\end{tabular}}
\end{table*}

\subsection{Synthetic Time-Series Generation}
\label{app:synthetic_data}

Following ChatTS \citep{xie2024chatts}, we synthesize base signals as $x_i=g(i)+s(i)+\epsilon(i)+\ell(i)$ for $i=1,\ldots,L$, where $L$ is the sequence length and the four terms represent trend, seasonality, noise, and local events, respectively. Their parameters control component shape, magnitude, and temporal extent. Multivariate and recurrent variants introduce shared shapes, temporally aligned events, and repeated templates. Table~\ref{tab:synthetic_design} summarizes representative constructions and their controlled parameters; the corresponding metadata supports subsequent interaction annotation (Appendix~\ref{app:synthetic_annotation}).

\begin{table}[!htbp]
\centering
\caption{Representative mechanisms and controlled parameters for synthetic time-series generation.}
\label{tab:synthetic_design}
\small
\setlength{\tabcolsep}{5pt}
\setlength{\aboverulesep}{2pt}
\setlength{\belowrulesep}{2pt}
\renewcommand{\arraystretch}{0.8}

\begin{tabular}{@{}
    >{\raggedright\arraybackslash}m{2.35cm}
    >{\raggedright\arraybackslash}m{4.8cm}
    >{\raggedright\arraybackslash}m{4.55cm}
@{}}
\toprule
\textbf{Structure} &
\textbf{Generation mechanism} &
\textbf{Controlled parameters} \\
\midrule

\rowcolor{black!5}
\multicolumn{3}{@{}l}{\textit{\textbf{Component composition}}} \\
\addlinespace[2pt]
Trend $g$ &
e.g., constant, monotonic, or piecewise curves &
Direction, magnitude, turning points \\
\addlinespace[2pt]
Seasonality $s$ &
e.g., harmonic, square, or triangular waves &
Waveform, period, amplitude \\
\addlinespace[2pt]
Noise $\epsilon$ &
e.g., Gaussian noise with optional oscillations &
Noise magnitude relative to signal scale \\
\addlinespace[2pt]
Local events $\ell$ &
e.g., spikes, dips, excursions, and level changes &
Type, onset, duration, magnitude \\
\midrule

\rowcolor{black!5}
\multicolumn{3}{@{}l}{\textit{\textbf{Multivariate composition}}} \\
\addlinespace[2pt]
Shared shape &
Shared shape with trend, seasonality, and noise &
Prototype perturbation, mixing weights, scale, offset \\
\addlinespace[2pt]
Aligned events &
Aligned events with trend, seasonality, and noise &
Timing jitter, event type, width, amplitude \\
\midrule

\rowcolor{black!5}
\multicolumn{3}{@{}l}{\textit{\textbf{Pattern recurrence}}} \\
\addlinespace[2pt]
Repeated template &
Shifted and rescaled copies of a fixed waveform &
Occurrence count, spacing, amplitude, target channel \\
\bottomrule
\end{tabular}
\end{table}

\subsection{Real-World Interaction Annotation Pipeline}
\label{app:annotation_pipeline}

\noindent\textbf{Model Configuration.}
We use GPT-5.6-Terra (\texttt{gpt-5.6-terra}) with
\texttt{xhigh} reasoning effort as LLM1 for scenario design,
temporal grounding, interaction generation, and reannotation.
DeepSeek-V4-Pro (\texttt{deepseek-v4-pro}) serves as LLM2 for
independent interaction verification, with a sampling temperature
of $0.7$. The two models are assigned distinct roles to separate
interaction construction from quality assessment during annotation.
Algorithm~\ref{alg:data_annotation} summarizes the complete annotation workflow, and the corresponding prompt templates are provided in Appendix~\ref{app:annotation_prompts}.

\noindent\textbf{Repeated Verification.}
We perform $n_v=3$ independent verification runs for each time series in every round. In each run, LLM2 evaluates the complete interaction set and returns a \texttt{PASS} or \texttt{FAIL} verdict, a confidence score, and an explanation for each failed interaction. The verification criteria cover temporal grounding, task consistency, response timing, future-information leakage, and unsupported claims. We aggregate the three independent judgments by majority vote to improve the robustness of the verification process. An interaction is selected for reannotation when at least $m=\lfloor n_v/2\rfloor+1$ runs return \texttt{FAIL}. Under our setting, this corresponds to at least two failed judgments. Only interactions meeting this criterion are passed to the subsequent reannotation stage.

\begin{algorithm}[t]
\caption{Streaming Interaction Annotation and Curation}
\label{alg:data_annotation}
\small
\begin{algorithmic}[1]
\algrenewcommand\algorithmicindent{1em}
\algrenewcommand\algorithmicrequire{\textbf{Input:}}
\algrenewcommand\algorithmicensure{\textbf{Output:}}
\algrenewcommand{\algorithmiccomment}[1]{\hfill\texttt{//}\ #1}

\Require series/metadata $\{(\mathbf{X}_i,\mathcal{M}_i)\}$,
templates/rules $\mathcal{T}$, verification count $n_v$,
confidence threshold $\tau_{\mathrm{conf}}$
\Ensure curated dataset $\mathcal{D}$

\State $\mathcal{D}\gets\emptyset$;
$m\gets\lfloor n_v/2\rfloor+1$
\For{each $(\mathbf{X}_i,\mathcal{M}_i)$}

    \Statex $\triangleright$ \textbf{Phase 1: Interaction Construction}
    \Comment{$\mathcal{S}$: scenarios; $\mathcal{G}$: evidence}
    \State $(\mathcal{S},\mathcal{G})\gets
    \Call{LLM1-DesignAndGround}{
        \mathbf{X}_i,\mathcal{M}_i,\mathcal{T}}$
    \State $\mathcal{A}=(a_j)_j\gets
    \Call{LLM1-Generate}{
        \mathbf{X}_i,\mathcal{S},\mathcal{G},\mathcal{T}}$

    \Repeat
        \Statex $\triangleright$ \textbf{Phase 2: Repeated Verification}
        \Comment{$v$: verdict; $c$: confidence; $f$: feedback}
        \For{verification run $k=1,\ldots,n_v$}
            \State $\mathcal{V}^{(k)}
            =\{(v_j^{(k)},c_j^{(k)},f_j^{(k)})\}_j
            \gets
            \Call{LLM2-Verify}{
                \mathbf{X}_i,\mathcal{S},\mathcal{G},\mathcal{A}}$
        \EndFor
        \State $\mathcal{F}\gets
        \left\{j:
            \sum_{k=1}^{n_v}
            \mathbf{1}[v_j^{(k)}=\texttt{FAIL}]
            \geq m
        \right\}$

        \If{$\mathcal{F}\neq\emptyset$}
            \Statex $\triangleright$ \textbf{Phase 3: Selective Reannotation}
            \State $\mathcal{A}_{\mathcal{F}}\gets
            \Call{LLM1-Reannotate}{
                \mathbf{X}_i,\mathcal{S},\mathcal{G},
                \mathcal{A},
                \{\mathcal{V}^{(k)}\}_{k=1}^{n_v},
                \mathcal{F}}$
        \EndIf
    \Until{$\mathcal{F}=\emptyset$}

    \Statex $\triangleright$ \textbf{Phase 4: Final Curation}
    \For{$j\in\Call{RuleFilter}{\mathcal{A},\mathcal{T}}$}
        \State $a'_j\gets a_j$
        \If{$\Call{AggregateConfidence}{
            \{c_j^{(k)}\}_{k=1}^{n_v}}
            <\tau_{\mathrm{conf}}$}
            \State $a'_j\gets
            \Call{HumanReview}{
                \mathbf{X}_i,\mathcal{S},\mathcal{G},a_j}$
            \Comment{$\emptyset$ if rejected}
        \EndIf
        \State $\mathcal{D}\gets
        \mathcal{D}\cup
        \bigl(\{a'_j\}\setminus\{\emptyset\}\bigr)$
    \EndFor
\EndFor
\State \Return $\mathcal{D}$

\end{algorithmic}
\end{algorithm}

\noindent\textbf{Failed-Interaction Reannotation.} LLM1 receives the time series, interaction scenarios and temporal evidence, the interaction set, and verification feedback. It regenerates only the interactions selected for reannotation, while all others remain unchanged. Each revised interaction may update the response position, user input, and response content. The updated interaction set is then evaluated through $n_v$ fresh LLM2 runs, with votes recomputed independently at each round rather than accumulated across rounds. This process continues until no interaction receives a majority of \texttt{FAIL} verdicts. With $R$ reannotation rounds, the procedure requires $n_v(R+1)$ verification calls during the full process.

\noindent\textbf{Final Curation.}
After LLM-based verification is completed, we first apply rule-based checks to enforce annotation structure and task-specific constraints. Samples that pass these checks are then evaluated using a confidence threshold $\tau_{\mathrm{conf}}$ based on the verification confidence scores. High-confidence samples are retained automatically, while samples with confidence below $\tau_{\mathrm{conf}}$ are sent for manual review and are subsequently accepted, corrected, or rejected.

\subsection{Synthetic Interaction Annotation}
\label{app:synthetic_annotation}

We generate synthetic interactions from temporal metadata using task-specific templates, with the component--task mappings summarized in Table~\ref{tab:interaction_evidence}. (1) \emph{Task assignment:} we select relevant component attributes and variables to construct current-state queries, historical reviews, and monitoring instructions, while other components remain as background.  (2) \emph{Response generation:} templates determine response timing and content. IQA answers describe the requested evidence; PIF notifications follow matching events or reporting schedules; PTW warnings target salient events or injected data corruption; and UGA adapts response behavior according to user instructions or feedback, while also learning user-provided concepts and identifying their subsequent occurrences. (3) \emph{Multi-turn composition:} we construct multi-turn interactions by combining compatible single-turn interactions from the same time series in chronological order, preserving user instructions and responses. Each multi-turn interaction contains at most ten model responses along the time-series stream.

\begin{table}[t]
\centering
\caption{
Task-specific temporal evidence and interaction construction patterns for synthetic annotation.
$\checkmark$ denotes eligible evidence and -- unused evidence for each task.
Eligible sources may be used independently or jointly, while other components remain as background throughout the sequence.
}
\label{tab:interaction_evidence}

\small
\setlength{\tabcolsep}{2.5pt}
\setlength{\aboverulesep}{0.7pt}
\setlength{\belowrulesep}{0.7pt}
\renewcommand{\arraystretch}{0.1}
\renewcommand{\tabularxcolumn}[1]{m{#1}}

\begin{tabularx}{\textwidth}{
    >{\centering\arraybackslash}m{0.08\textwidth}
    >{\centering\arraybackslash}m{0.21\textwidth}
    ccccc
    >{\raggedright\arraybackslash}X
}
\toprule
\multirow{2}{=}[-0.5\baselineskip]{\centering\textbf{Level}}
& \multirow{2}{=}[-0.5\baselineskip]{\centering\textbf{Task}}
& \multicolumn{5}{c}{\textbf{Temporal Evidence}}
& \multirow{2}{=}[-0.5\baselineskip]{\centering\textbf{Interaction Pattern}} \\
\cmidrule(lr){3-7}
& &
\textbf{Trend} &
\makecell{\textbf{Season-}\\[-1pt]\textbf{ality}} &
\makecell{\textbf{Cross-}\\[-1pt]\textbf{var.}} &
\textbf{Event} &
\textbf{Corrupt.} &
\\
\midrule

\multirow{2}{=}[-0.5\baselineskip]{\centering IQA}
& \makecell{Current State\\[-1pt]QA}
& $\checkmark$ & $\checkmark$ & $\checkmark$ & $\checkmark$ & --
& Attribute query $\rightarrow$ current-state response \\
& \makecell{Historical\\[-1pt]Review}
& $\checkmark$ & -- & $\checkmark$ & $\checkmark$ & --
& Past interval $\rightarrow$ trend/event review \\
\midrule

\multirow{2}{=}[-0.5\baselineskip]{\centering PIF}
& \makecell{Event\\[-1pt]Watch}
& -- & -- & $\checkmark$ & $\checkmark$ & --
& Monitoring instruction $\rightarrow$ event-triggered report \\
& \makecell{Repeated\\[-1pt]Reporting}
& -- & -- & $\checkmark$ & -- & --
& Reporting schedule $\rightarrow$ periodic notification \\
\midrule

\multirow{2}{=}[-0.5\baselineskip]{\centering PTW}
& \makecell{Pattern\\[-1pt]Warning}
& -- & -- & $\checkmark$ & $\checkmark$ & --
& Salient event $\rightarrow$ proactive warning \\
& \makecell{Data-Quality\\[-1pt]Warning}
& -- & -- & $\checkmark$ & -- & $\checkmark$
& Corrupted interval $\rightarrow$ data-quality warning \\
\midrule

\multirow{2}{=}[-0.5\baselineskip]{\centering UGA}
& \makecell{Feedback-Based\\[-1pt]Monitoring}
& -- & -- & $\checkmark$ & $\checkmark$ & --
& Feedback-adjusted criterion $\rightarrow$ selective alert \\
& \makecell{Demonstration-Guided\\[-1pt]Concept Learning}
& -- & -- & $\checkmark$ & $\checkmark$ & --
& New example $\rightarrow$ recurrence-triggered report \\
\bottomrule
\end{tabularx}
\end{table}

\subsection{Annotation Prompt Templates}
\label{app:annotation_prompts}

We provide the prompt templates for scenario design and temporal grounding,
interaction generation, verification, and regeneration. Instructions and
input placeholders are retained; detailed examples and output schemas are
abbreviated as \texttt{\{Example Return\}} and
\texttt{\{Output Format\}}, respectively.

\refstepcounter{annotationprompt}
\begin{tcolorbox}[
    enhanced,
    breakable,
    colback=white,
    colframe=PromptBorder,
    colbacktitle=PromptTitle,
    coltitle=white,
    boxrule=0.3mm,
    rounded corners,
    title={Prompt 01: Scenario Design and Component Grounding},
    fonttitle=\bfseries,
    left=2mm, right=2mm, top=2mm, bottom=2mm,
    before upper={
        \setlength{\parindent}{0pt}
        \raggedright
        \setstretch{1.0}
    }
]
\label{prompt:scenario_design}

\textbf{Role}\\[3pt]
You are a time-series interaction annotator responsible for designing interactive scenarios and identifying the temporal evidence that supports each scenario.

\par\medskip
\textbf{Task}
\begin{itemize}[
    leftmargin=1.25em,
    itemsep=2pt,
    parsep=0pt,
    topsep=3pt,
    after=\vspace{3pt}
]
    \item From \texttt{Instant Query Answering (IQA)},
    \texttt{Persistent Instruction Following (PIF)},
    \texttt{Proactive Temporal Warning (PTW)}, and
    \texttt{User-Guided Adaptation (UGA)}, select \textbf{the most suitable task
    types} for the provided time series and design
    \textbf{one or more appropriate interaction scenarios}.

    \item Multiple scenarios may belong to the same task type.

    \item For each scenario:
    \begin{itemize}[
        leftmargin=1.25em,
        itemsep=3pt,
        parsep=0pt,
        topsep=3pt
    ]
        \item Describe the interaction that could naturally occur
        based on the observed temporal behavior.
        \item Identify \textbf{temporal evidence instances} in the
        time series that support the scenario.
    \end{itemize}

    \item One scenario may contain multiple temporal evidence instances
    occurring at different intervals.

    \item Only construct scenarios and temporal evidence that are
    clearly supported by the data.

    \item Do not infer unsupported causes or explanations.
\end{itemize}

\textbf{Example}\\[3pt]
\texttt{\{example\_return\}}

\par\medskip
\textbf{Variable Descriptions}\\[3pt]
\texttt{\{variable\_descriptions\}}

\par\medskip
\textbf{Time-Series Data}\\[3pt]
\texttt{\{time\_series\_data\}}

\par\medskip
\textbf{Output Format}\\[3pt]
\texttt{\{output\_format\}}
\end{tcolorbox}

\refstepcounter{annotationprompt}
\begin{tcolorbox}[
    enhanced,
    breakable,
    colback=white,
    colframe=PromptBorder,
    colbacktitle=PromptTitle,
    coltitle=white,
    boxrule=0.3mm,
    rounded corners,
    title={Prompt 02: Scenario-Grounded Interaction Generation},
    fonttitle=\bfseries,
    left=2mm, right=2mm, top=2mm, bottom=2mm,
    before upper={
        \setlength{\parindent}{0pt}
        \raggedright
        \setstretch{1.0}
    }
]
\label{prompt:interaction_generation}

\textbf{Role}\\[3pt]
You are a time-series interaction annotator responsible for generating user-assistant interactions based on given interaction scenarios and their temporal evidence.

\par\medskip
\textbf{Scenarios}\\[3pt]
The following JSON array contains all scenarios for one time-series sample. Process the scenarios in the listed order.
\\[3pt]
\texttt{\{scenarios\}}

\par\medskip
\textbf{Task}\\[3pt]
Generate user-assistant interaction samples that naturally instantiate the given scenarios.

\begin{itemize}[
    leftmargin=1.25em,
    itemsep=2pt,
    parsep=0pt,
    topsep=3pt,
    after=\vspace{3pt}
]

    \item Generate one interaction for each
    appropriate \textbf{response-triggering temporal evidence}, in the same order as the supplied temporal evidences.

    \item Select an appropriate \texttt{query\_point} where sufficient evidence is available to support the response.

    \item Use only observations at or before \texttt{query\_point}.

    \item Ground each interaction in the corresponding temporal
    evidence and scenario.

    \item Keep the interactions \textbf{concise and consistent} with the
    corresponding scenario and task type.

    \item If no user input occurs at the response point, set
    \texttt{question} to an empty string.

    \item Do not use future observations or infer unsupported
    causes or explanations.
\end{itemize}

\textbf{Example}\\[3pt]
\texttt{\{example\_return\}}

\par\medskip
\textbf{Variable Descriptions}\\[3pt]
\texttt{\{variable\_descriptions\}}

\par\medskip
\textbf{Time-Series Data}\\[3pt]
\texttt{\{time\_series\_data\}}

\par\medskip
\textbf{Output Format}\\[3pt]
\texttt{\{output\_format\}}
\end{tcolorbox}

\refstepcounter{annotationprompt}
\begin{tcolorbox}[
    enhanced,
    breakable,
    colback=white,
    colframe=PromptBorder,
    colbacktitle=PromptTitle,
    coltitle=white,
    boxrule=0.3mm,
    rounded corners,
    title={Prompt 03: Evidence-Grounded Interaction Verification},
    fonttitle=\bfseries,
    left=2mm, right=2mm, top=2mm, bottom=2mm,
    before upper={
        \setlength{\parindent}{0pt}
        \raggedright
        \setstretch{1.0}
    }
]
\label{prompt:interaction_verification}

\textbf{Role}\\[3pt]
You are a time-series interaction verifier responsible for evaluating generated user-assistant interactions against the given scenarios and temporal evidence.

\par\medskip
\textbf{Scenarios}\\[3pt]
\texttt{\{scenarios\}}

\par\medskip
\textbf{Generated Interactions}\\[3pt]
\texttt{\{generated\_interactions\}}

\par\medskip
\textbf{Task}\\[3pt]
Evaluate whether each generated interaction is valid with respect to its corresponding scenario and temporal evidence.

\begin{itemize}[
    leftmargin=1.25em,
    itemsep=2pt,
    parsep=0pt,
    topsep=3pt,
    after=\vspace{3pt}
]
    \item Check whether each interaction is \textbf{consistent with the
    corresponding scenario, task type, and temporal evidence}.

    \item Check whether each \texttt{query\_point} is appropriate and
    whether sufficient evidence is available at that point.

    \item Check whether each question and answer accurately reflects
    the observed time-series behavior.

    \item Check whether any interaction uses \textbf{future observations}
    or introduces unsupported claims.

    \item Return \texttt{PASS} only if the interaction is valid.

    \item Report \texttt{confidence} from 0.0 to 1.0 for each interaction verdict.
\end{itemize}

\textbf{Example}\\[3pt]
\texttt{\{example\_return\}}

\par\medskip
\textbf{Variable Descriptions}\\[3pt]
\texttt{\{variable\_descriptions\}}

\par\medskip
\textbf{Time-Series Data}\\[3pt]
\texttt{\{time\_series\_data\}}

\par\medskip
\textbf{Output Format}\\[3pt]
\texttt{\{output\_format\}}
\end{tcolorbox}

\refstepcounter{annotationprompt}
\begin{tcolorbox}[
    enhanced,
    breakable,
    colback=white,
    colframe=PromptBorder,
    colbacktitle=PromptTitle,
    coltitle=white,
    boxrule=0.3mm,
    rounded corners,
    title={Prompt 04: Verification-Guided Interaction Regeneration},
    fonttitle=\bfseries,
    left=2mm, right=2mm, top=2mm, bottom=2mm,
    before upper={
        \setlength{\parindent}{0pt}
        \raggedright
        \setstretch{1.0}
    }
]
\label{prompt:interaction_regeneration}

\textbf{Role}\\[3pt]
You are a time-series interaction annotator responsible for regenerating user-assistant interactions that did not pass verification based on given interaction scenarios and their temporal evidence.

\par\medskip
\textbf{Scenarios}\\[3pt]
The following JSON array contains the scenarios corresponding to the failed interactions.
\\[3pt]
\texttt{\{scenarios\}}

\par\medskip
\textbf{Failed Verification Result}\\[3pt]
\texttt{\{failed\_verification\_result\}}

\par\medskip
\textbf{Task}\\[3pt]
Regenerate user-assistant interaction samples that instantiate the given scenarios while correcting the problems identified during verification.

\begin{itemize}[
    leftmargin=1.25em,
    itemsep=2pt,
    parsep=0pt,
    topsep=3pt,
    after=\vspace{3pt}
]

    \item Generate new interactions rather than reproducing the failed interactions.

    \item Correct all problems identified in the corresponding
    \textbf{verification result}.

    \item Generate one interaction for each appropriate
    \textbf{response-triggering temporal evidence}, in the same order as the supplied temporal evidences.

    \item Select an appropriate \texttt{query\_point} where sufficient evidence is available to support the response.

    \item Use only observations at or before \texttt{query\_point}.

    \item Ground each interaction in the corresponding temporal evidence and scenario.

    \item Keep the interactions concise and consistent with the corresponding scenario and task type.

    \item If no user input occurs at the response point, set
    \texttt{question} to an empty string.

    \item Do not use future observations or infer unsupported causes or explanations.
\end{itemize}

\textbf{Variable Descriptions}\\[3pt]
\texttt{\{variable\_descriptions\}}

\par\medskip
\textbf{Time-Series Data}\\[3pt]
\texttt{\{time\_series\_data\}}

\par\medskip
\textbf{Output Format}\\[3pt]
\texttt{\{output\_format\}}
\end{tcolorbox}

\section{Additional Experiments}
\label{app:additional_experiments}

\begin{figure}[t]
    \centering
    \includegraphics[width=\linewidth]{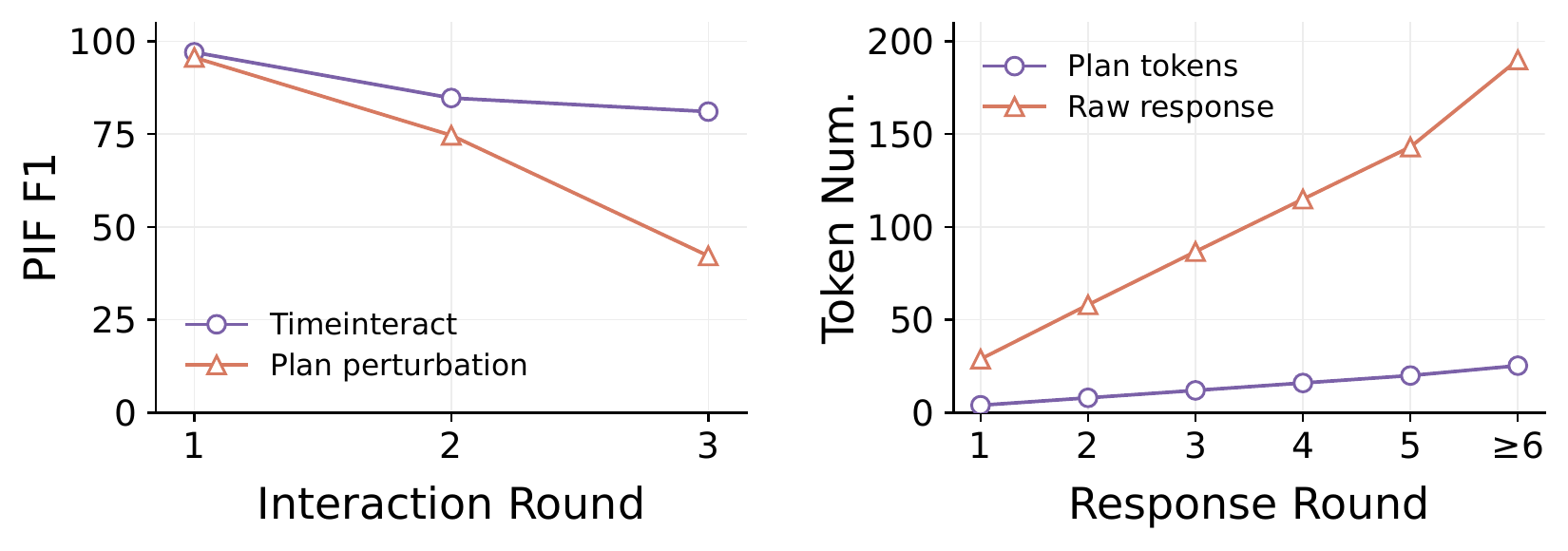}
    \caption{Analysis of Plan Tokens. Left: PIF F1 across rounds with original and perturbed Plan embeddings. Right: mean history token counts for Plan Tokens and full response history.}
    \label{fig:additional_plan_analysis}
    \vspace{-4mm}

\end{figure}

\noindent\textbf{Analysis of Plan Tokens.}
Figure~\ref{fig:additional_plan_analysis} examines the role of Plan Tokens in subsequent response triggering and history compression. On PIF (left), perturbing Plan embeddings widens the F1 gap across the first three interaction rounds, with third-round F1 dropping from 81 to 42. For history compression (right), we compare retaining only Plan Tokens with retaining the full response history. In the pooled $\geq6$ bin, mean history length drops from 190 to 25 tokens, a reduction of 86.8\%. These results show that Plan Tokens serve as a compact history representation for subsequent response triggering.

\noindent\textbf{Out-of-Domain Evaluation.}
We further evaluate response triggering on an additional OOD dataset containing 200 interaction episodes from eight real-world domains, with 100 episodes from each annotated single-turn and multi-turn subset. All eight models are evaluated on the same sampled episodes, each containing at most six variables. We retain the model settings and response-triggering metrics used in the main evaluation and report precision, recall, F1, and NQ-F1.

As shown in Table~\ref{tab:ood_triggering_results}, \method achieves the highest precision, F1, and NQ-F1 in both single-turn and multi-turn OOD settings. In the multi-turn setting, its No-Query F1 reaches 37.40\%, outperforming the strongest baseline by 21.87 percentage points. These results indicate that \method generalizes well to OOD streams while maintaining effective response triggering.

\begin{table}[ht]
\centering
\caption{
Response-triggering performance on the OOD test set. All scores are percentages. Bold and underlined values indicate the best and second-best results in each column, respectively.}
\label{tab:ood_triggering_results}
\Large
\setlength{\tabcolsep}{3pt}
\renewcommand{\arraystretch}{1}
\resizebox{\textwidth}{!}{\begin{tabular}{
@{}l c c c
*{8}{>{\centering\arraybackslash}m{3.3em}}
@{}
}
\toprule
\multirow{2.5}{*}{\textbf{Model}}
& \multirow{2.5}{*}{\textbf{Size}}
& \multirow{2.5}{*}{\makecell[c]{\textbf{Model}\\[-1pt]\textbf{Input}}}
& \multirow{2.5}{*}{\makecell[c]{\textbf{Native}\\[-1pt]\textbf{Streaming}}}
& \multicolumn{4}{c}{\textbf{Single-turn}}
& \multicolumn{4}{c}{\textbf{Multi-turn}} \\
\cmidrule(lr){5-8}
\cmidrule(lr){9-12}
& & & &
\textbf{P} &
\textbf{R} &
\textbf{F1} &
\mbox{\textbf{NQ-F1}} &
\textbf{P} &
\textbf{R} &
\textbf{F1} &
\mbox{\textbf{NQ-F1}} \\
\midrule

\multicolumn{12}
{@{}>{\columncolor{headergreen}[0pt][0pt]}l@{}}
{\textit{\textbf{General LLMs}}} \\
Qwen2.5-7B-Instruct
& 7B & Text & \ding{55}
& \underline{55.86} & 43.78 & \underline{49.09} & \underline{26.00}
& \underline{82.67} & 65.23 & \underline{72.93} & 9.60 \\
Mistral-7B-Instruct-v0.3
& 7B & Text & \ding{55}
& 16.31 & 57.30 & 25.39 & 13.94
& 35.94 & 60.94 & 45.22 & 7.96 \\
Qwen3-14B
& 14B & Text & \ding{55}
& 22.11 & \underline{71.35} & 33.76 & 20.03
& 36.63 & 72.27 & 48.62 & \underline{15.53} \\

\multicolumn{12}
{@{}>{\columncolor{headergreen}[0pt][0pt]}l@{}}
{\textit{\textbf{Vision-Language Models}}} \\
Qwen2.5-VL-7B
& 7B & Figure+Text & \ding{55}
& 10.07 & 23.24 & 14.05 & 8.53
& 17.68 & 22.66 & 19.86 & 9.68 \\
InternVL3.5-8B
& 8B & Figure+Text & \ding{55}
& 9.92 & \textbf{100.00} & 18.05 & 11.58
& 13.95 & \textbf{100.00} & 24.49 & 9.62 \\

\multicolumn{12}
{@{}>{\columncolor{headergreen}[0pt][0pt]}l@{}}
{\textit{\textbf{Time-Series Language Models}}} \\
ChatTS
& 14B & TS+Text & \ding{55}
& 10.85 & 60.00 & 18.38 & 13.84
& 11.07 & 32.81 & 16.55 & 9.77 \\
TimeOmni-1
& 7B & TS+Text & \ding{55}
& 51.25 & 22.16 & 30.94 & 11.39
& 80.21 & 30.08 & 43.75 & 5.66 \\

\multicolumn{12}
{@{}>{\columncolor{headergreen}[0pt][0pt]}l@{}}
{\textit{\textbf{TS Interaction Model}}} \\
\textbf{\method}
& 4B & Streaming TS+Text & \ding{51}
& \textbf{69.01} & 63.78 & \textbf{66.29} & \textbf{41.75}
& \textbf{92.42} & \underline{76.17} & \textbf{83.51} & \textbf{37.40} \\
\bottomrule
\end{tabular}}
\end{table}

\section{Broader Impact}
\label{appdix:broader_impact}

\noindent\textbf{Toward interactive time-series intelligence.}
To the best of our knowledge, this work is the first to introduce the concept of \emph{Time-Series Interaction}, extending time-series language models from offline question answering toward real-time interaction over continuously evolving streams. Rather than treating time series as static inputs that are analyzed only after they are fully observed, our formulation emphasizes continuous perception, response timing, persistent instructions, proactive interaction, and adaptation to user feedback. We hope this perspective can stimulate further exploration of interactive intelligence for time-series data, including new model architectures, learning paradigms, benchmarks, and evaluation protocols for continuously evolving environments.

\noindent\textbf{Potential for Edge Deployment.} The streaming design of \method shows strong potential for deployment on edge devices. Since observations are processed incrementally rather than repeatedly re-encoding the entire historical sequence, \method is naturally suited to continuously generated sensor streams. With further advances in lightweight model design, model compression, and efficient inference, such systems could potentially operate on wearable devices, mobile platforms, industrial controllers, and embedded monitoring systems, enabling low-latency interaction.

\noindent\textbf{Human-centered Time-Series Interaction.} \method places users at the center of time-series interaction by allowing them to continuously express their needs through queries, instructions, and feedback. Users can decide what information is relevant, how the analysis should be presented, and how the interaction behavior should change as their goals evolve. Rather than forcing users to adapt to a fixed analysis procedure, the system can follow and refine its behavior according to user intent throughout the stream. This provides a more flexible and natural way for people to interact with evolving time-series data and supports more effective human--AI collaboration.

\section{Limitations and Future Work}
\label{appdix:future_work}

Although \dataset provides large-scale time-series interaction data spanning both synthetic and real-world time series across multiple domains, the current interaction construction still relies primarily on synthetic scenarios. In practice, real-world settings involve much richer and more diverse interaction scenarios and user behaviors. Future work should therefore extend Time-Series Interaction to a broader range of real-world systems with native interactive capabilities, enabling the collection and study of more naturally occurring interaction scenarios and user behaviors. In addition, as Time-Series Interaction is still an emerging problem setting, establishing more standardized learning paradigms, benchmarks, and evaluation protocols will be important for advancing this direction.

\section{Case Studies}
\label{app:interaction_examples}

\begin{figure}[ht]
    \centering
    \includegraphics[width=\linewidth]{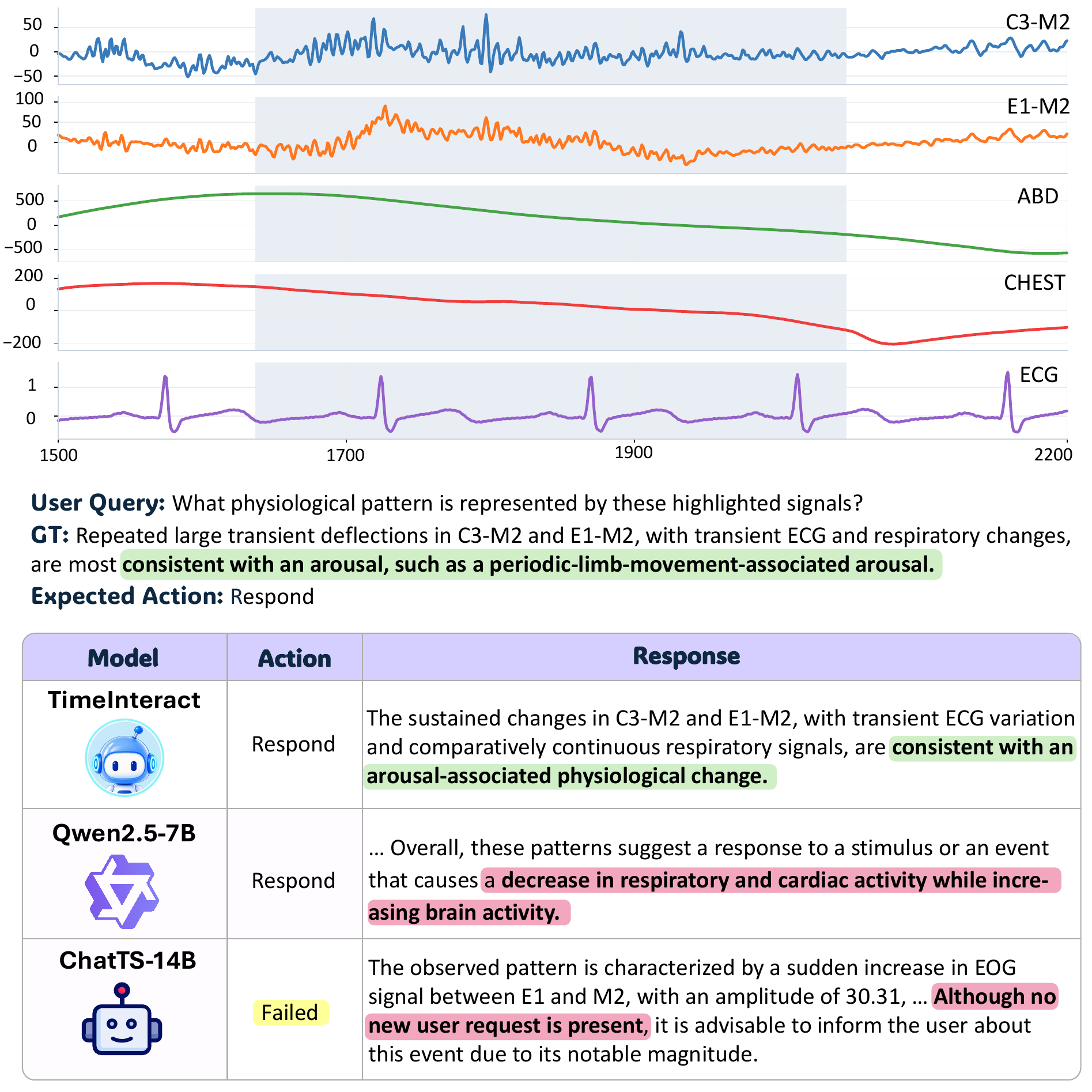}
    \caption{An example of IQA. \method interprets the highlighted physiological signals as arousal-related changes, consistent with the reference response.}
    \label{example1}
\end{figure}

\begin{figure}[t]
    \centering
    \includegraphics[width=\linewidth]{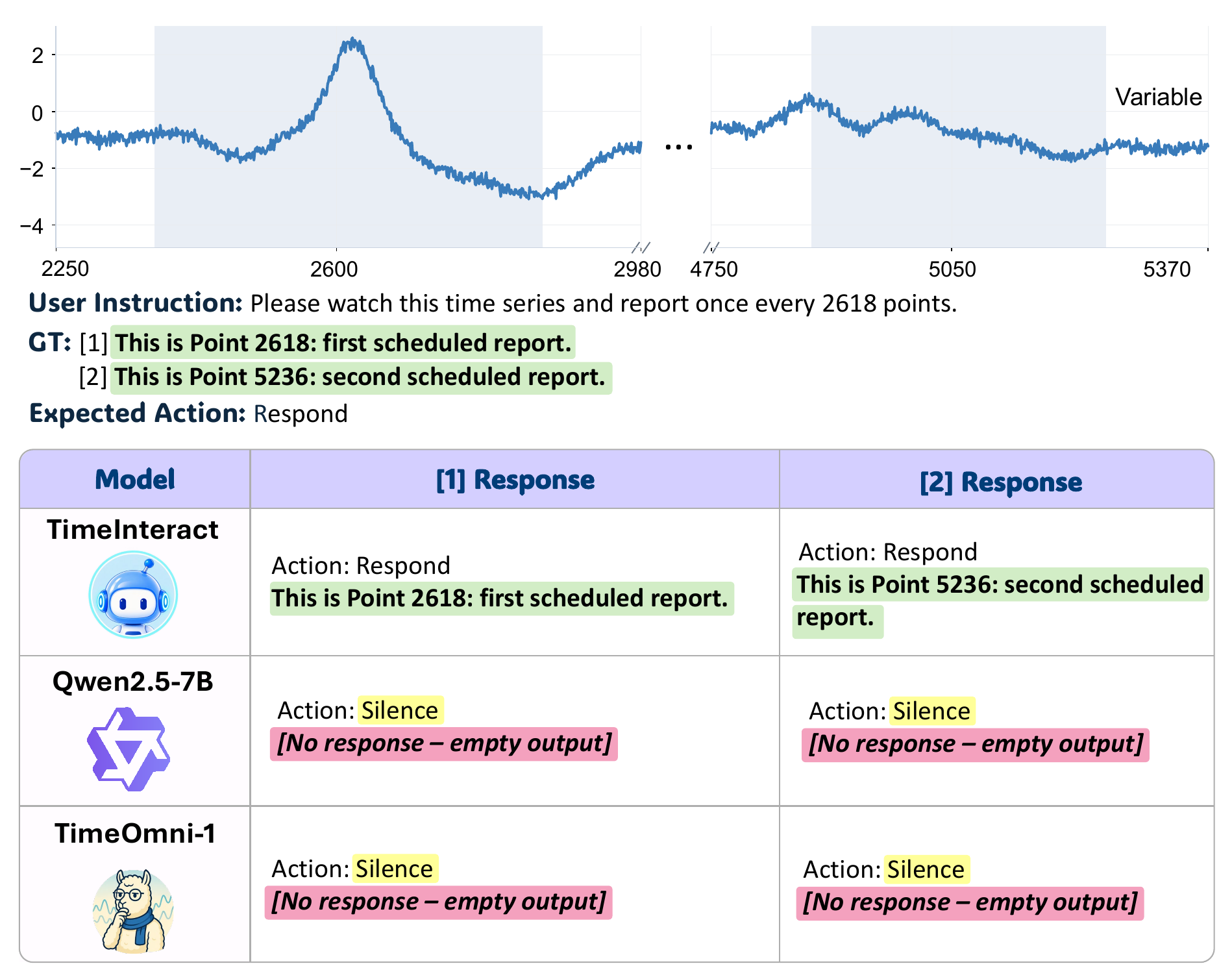}
    \caption{An example of PIF. \method follows the user's periodic-reporting instruction and responds at both scheduled points, while the baselines remain silent.}
    \label{example2}
\end{figure}

\begin{figure}[t]
    \centering
    \includegraphics[width=\linewidth]{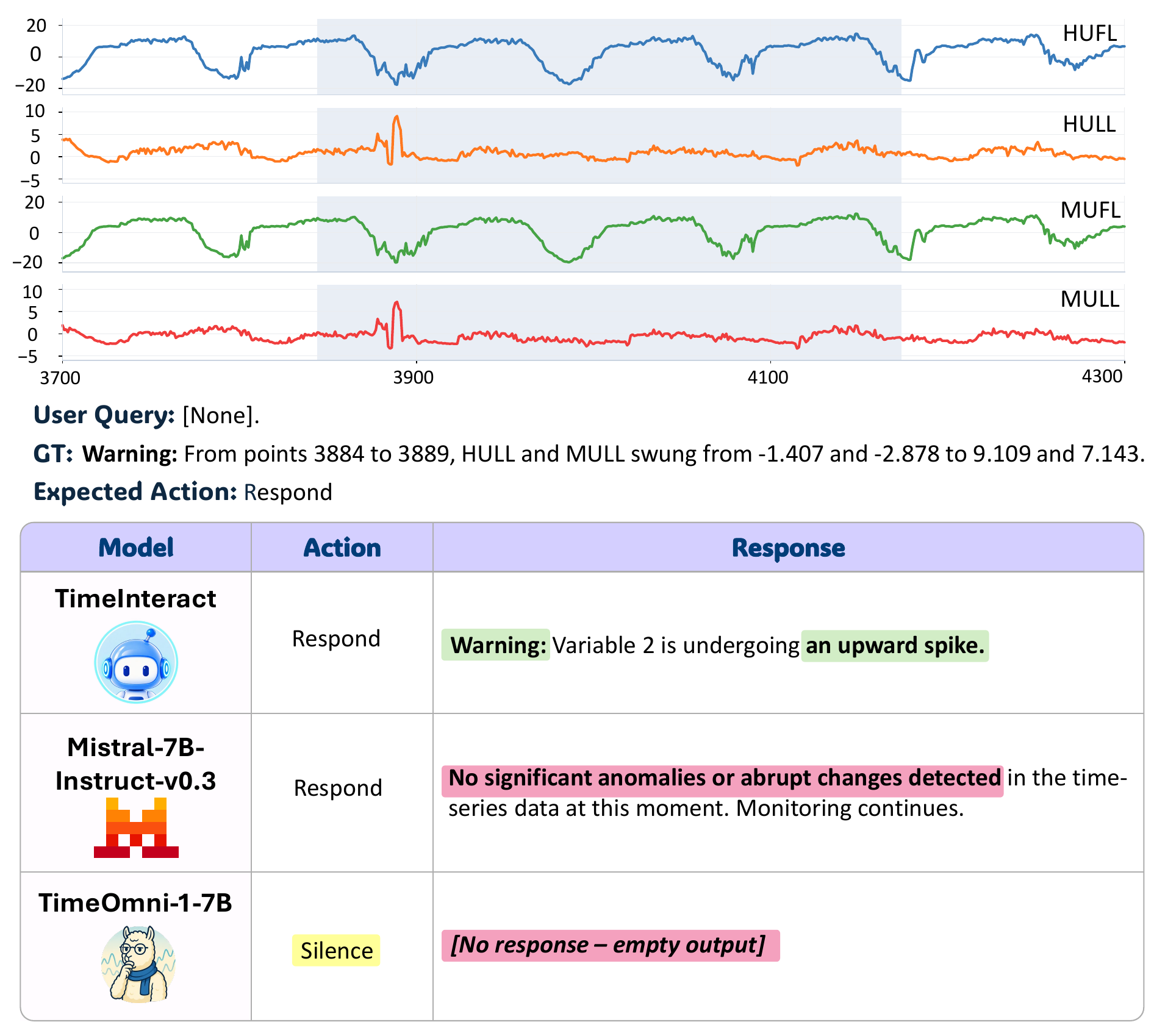}
    \caption{An example of PTW. \method warns of an upward spike without an explicit user query, while the baselines either report no anomaly or remain silent.}
    \label{example3}
\end{figure}

\begin{figure}[t]
    \centering
    \includegraphics[width=\linewidth]{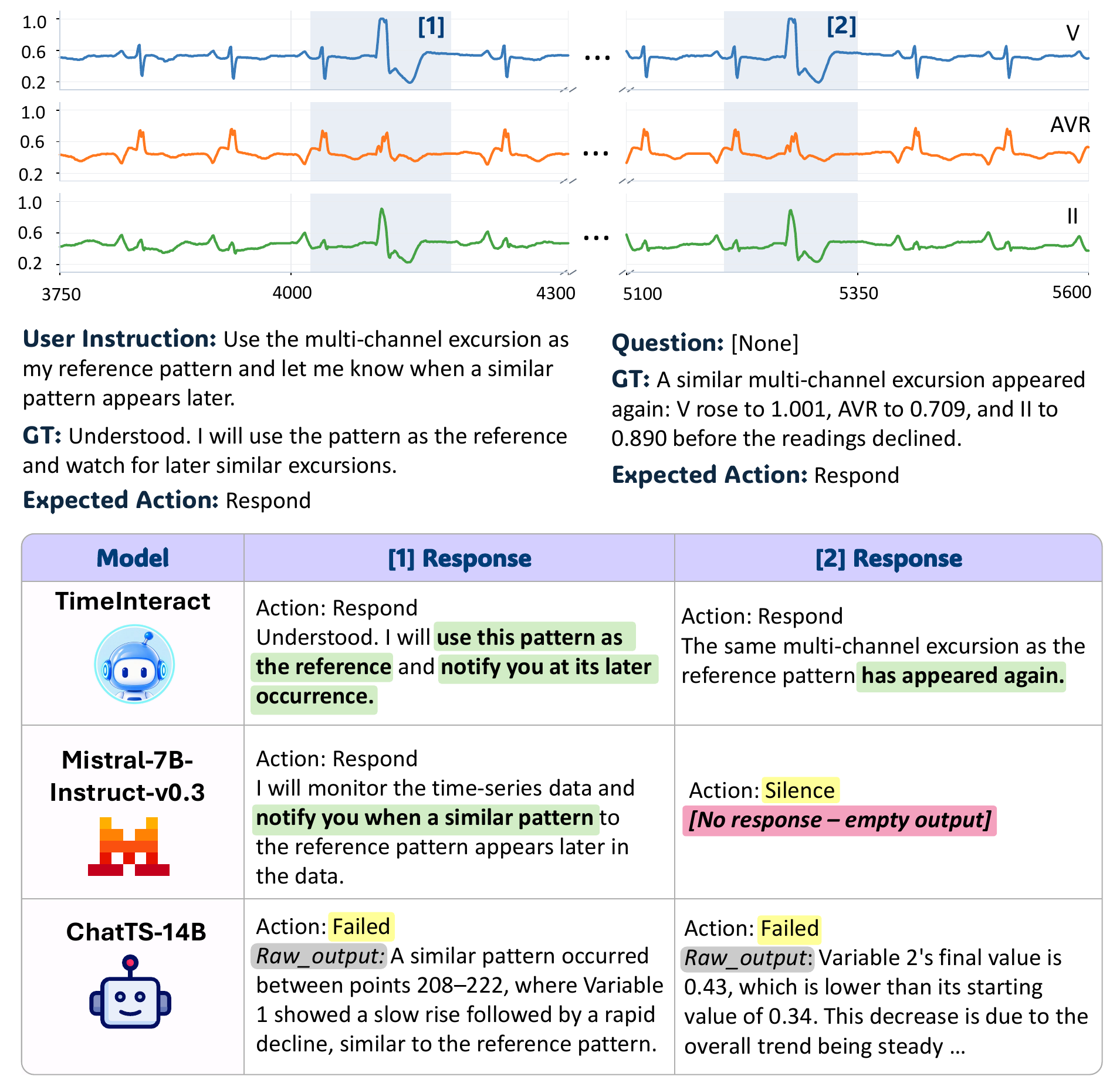}
    \caption{An example of UGA. \method adopts the demonstrated multichannel pattern as a reference and reports its later recurrence.}
    \label{example4}
\end{figure}

\begin{figure}[t]
    \centering
    \includegraphics[width=\linewidth]{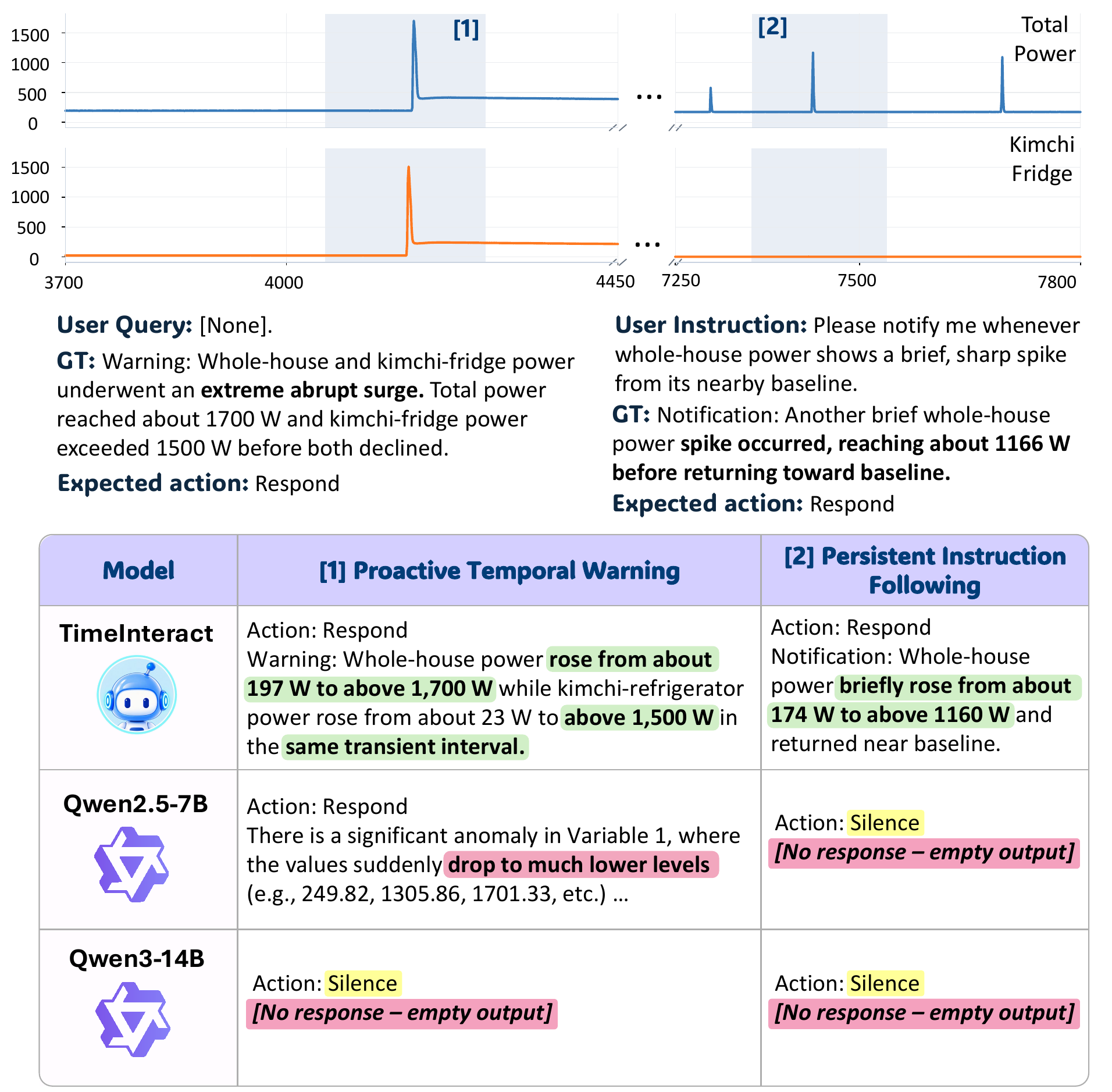}
    \caption{An example of multi-turn interaction combining PTW and PIF. \method first warns of a power surge, then follows a user instruction to report a later power spike.}
    \label{example5}
\end{figure}

\begin{figure}[t]
    \centering
    \includegraphics[width=\linewidth]{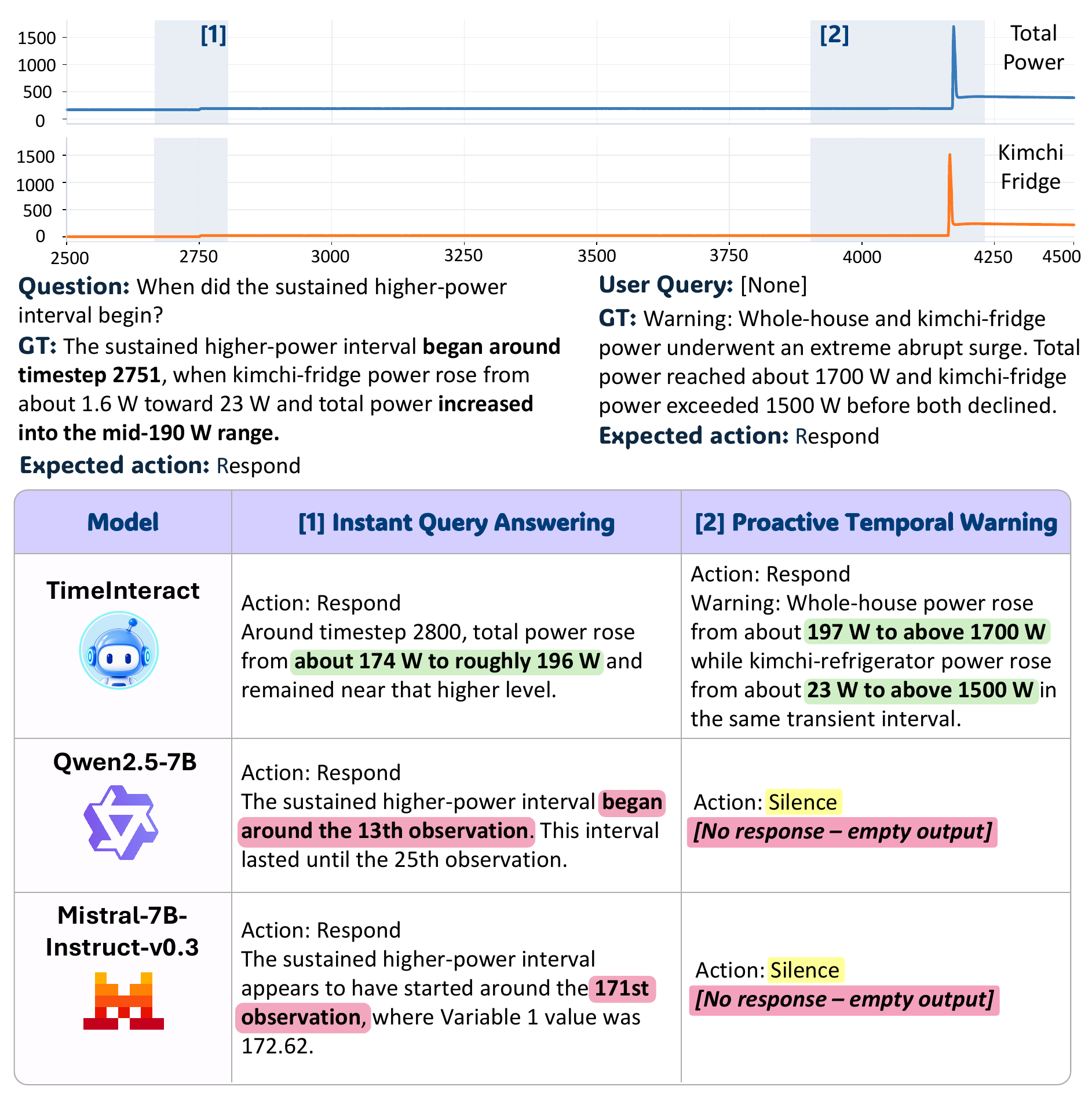}
    \caption{An example of multi-turn interaction combining IQA and PTW. \method estimates the onset of a higher-power interval and subsequently warns of a surge without a new query.}
    \label{example6}
\end{figure}

We provide six examples of streaming time-series interaction, comparing the response decisions and generated answers of \method with representative baselines. Each case includes time-series observations, user queries or instructions when present, and reference responses to contextualize the model outputs. Figures~\ref{example1}--\ref{example4} illustrate IQA, PIF, PTW, and UGA, respectively, covering query answering, instruction persistence, proactive warnings, and adaptation to user guidance. Figures~\ref{example5} and~\ref{example6} present two multi-turn scenarios combining PTW with PIF and IQA with PTW, respectively.

\end{document}